\documentclass[a4paper]{cas-dc}

\usepackage[authoryear]{natbib}
\usepackage{amsmath}
\usepackage{amssymb}
\usepackage{amsfonts}
\usepackage{hyphenat} 
\usepackage{subfigure}
\usepackage{bbding} 
\usepackage{multirow}
\def\tsc#1{\csdef{#1}{\textsc{\lowercase{#1}}\xspace}}
\tsc{WGM}
\tsc{QE}
\tsc{EP}
\tsc{PMS}
\tsc{BEC}
\tsc{DE}

\begin{document}
\let\WriteBookmarks\relax
\def\floatpagepagefraction{1}
\def\textpagefraction{.001}

\shorttitle{HiSup}

\shortauthors{Xu et~al.}

\title [mode = title]{
Accurate Polygonal Mapping of Buildings in Satellite Imagery 
}

\author[1,2]{Bowen Xu}
\fnmark[1]
\author[2]{Jiakun Xu}
\fnmark[1]
\author[2]{Nan Xue}
\cormark[1]
\author[1,2]{Gui-Song Xia}
\cormark[1]




\affiliation[1]{organization={State Key Lab. of LIESMARS, Wuhan University},
    city={Wuhan},
    postcode={430079}, 
    country={China}}
\affiliation[2]{organization={School of Computer Science, Wuhan University},
    city={Wuhan},
    postcode={430072}, 
    country={China}}

\cortext[cor1]{Corresponding authors.}

\fntext[fn1]{Equal contributions.}


\begin{abstract}
This paper studies the problem of polygonal mapping of buildings by tackling the issue of mask reversibility that leads to a notable performance gap between the predicted masks and polygons from the learning-based methods. We addressed such an issue by exploiting the hierarchical supervision (of bottom-level vertices, mid-level line segments and the high-level regional masks) and proposed a novel interaction mechanism of feature embedding sourced from different levels of supervision signals to obtain reversible building masks for polygonal mapping of buildings. As a result, we show that the learned reversible building masks take all the merits of the advances of deep convolutional neural networks for high-performing polygonal mapping of buildings. In the experiments, we evaluated our method on the two public benchmarks of AICrowd and Inria. On the AICrowd dataset, our proposed method obtains unanimous improvements on the metrics of AP, AP$^{\text{boundary}}$ and PoLiS. For the Inria dataset, our proposed method also obtains very competitive results on the metrics of IoU and Accuracy. The models and source code are available at \url{https://github.com/SarahwXU/HiSup}.
\end{abstract}

\maketitle

\section{Introduction}
Polygonal mapping of buildings, aiming at precisely extracting the footprint of buildings from high-resolution satellite imagery in the form of polygons, is a core and dynamic problem in photogrammetric computer vision and remote sensing~\citep{sohn2001extraction,zorzi2021polyworld}, and plays important roles in the Geographic Information Systems (GIS) for making up the basic feature classes~\citep{MAYER1999automatic}.

As in orthorectified satellite imagery, the footprints and roof outlines of buildings coincide under most circumstances, the mapping of buildings could be roughly equivalent to extracting building polygons~\citep{wang2022learning}. Thus, the problem of polygonal mapping of buildings in satellite imagery has usually been formulated in a multi-step paradigm~\citep{Girard2020polygonal,Li2021joint}, by (1) computing the rasterized binary masks of buildings from the input images, (2) vectorizing the binary masks into polygons with certain heuristic post-processing schemes, and (3) optionally simplifying the initial polygons to reduce the redundancy of vertices. In this paradigm, building segmentation, as a key sub-problem, has recently been significantly advanced by deep neural networks (DNNs)~\citep{mnih2013phd,Yuan2018learning}. Thereafter, the quality of polygonal mapping of buildings can be improved~\citep{Wei2020toward}
by leveraging the off-the-shelf ad-hoc algorithms such as Marching Squares~\citep{MarchingCubes} for polygon generation and the Douglas-Peucker~\citep{douglas1973al} for the shape simplification. 

\begin{figure}[!t]
\begin{center}
\subfigure[Frame-Field Learning~\citep{Girard2020polygonal}]{
\includegraphics[width=0.32\linewidth]{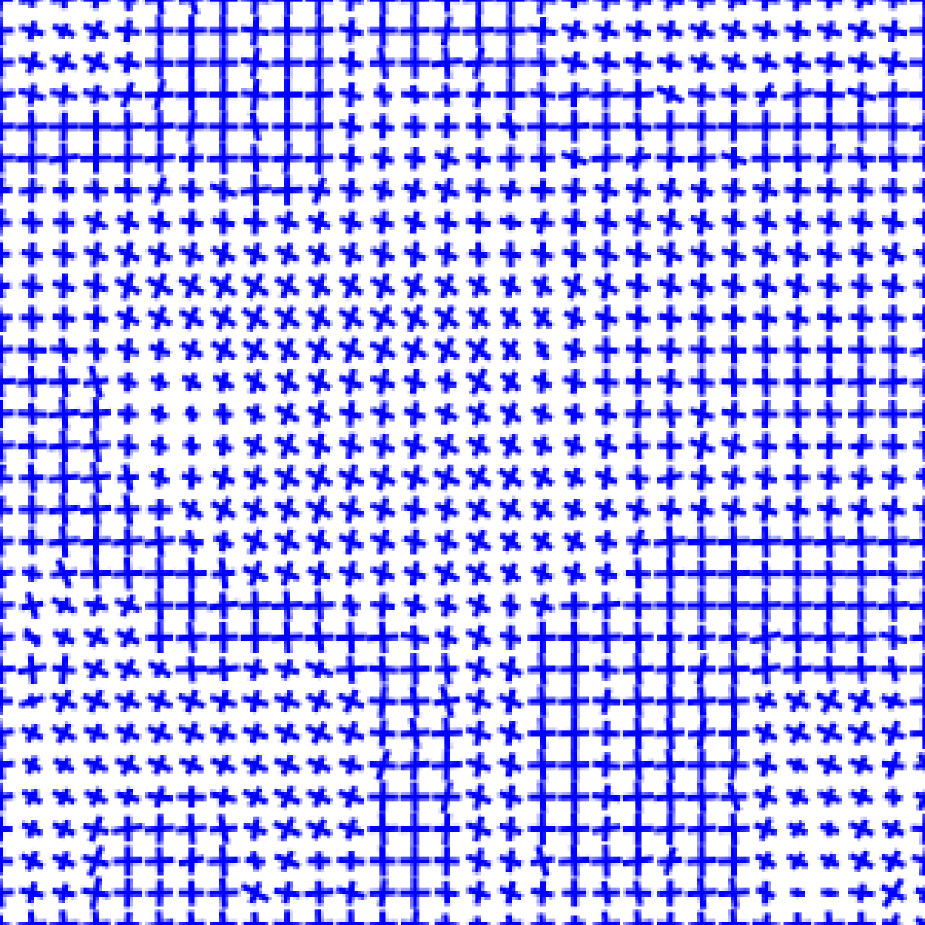}
\includegraphics[width=0.32\linewidth]{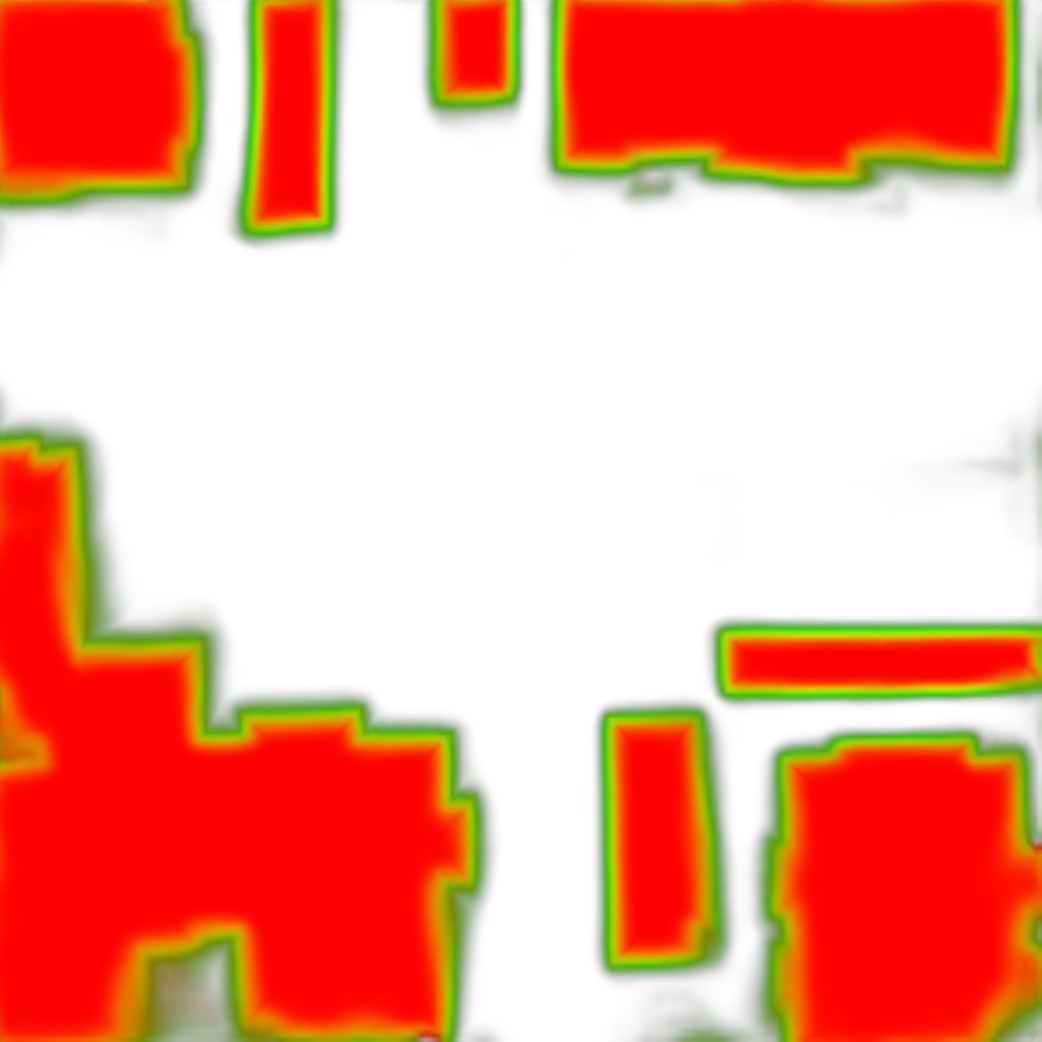}
\includegraphics[width=0.32\linewidth]{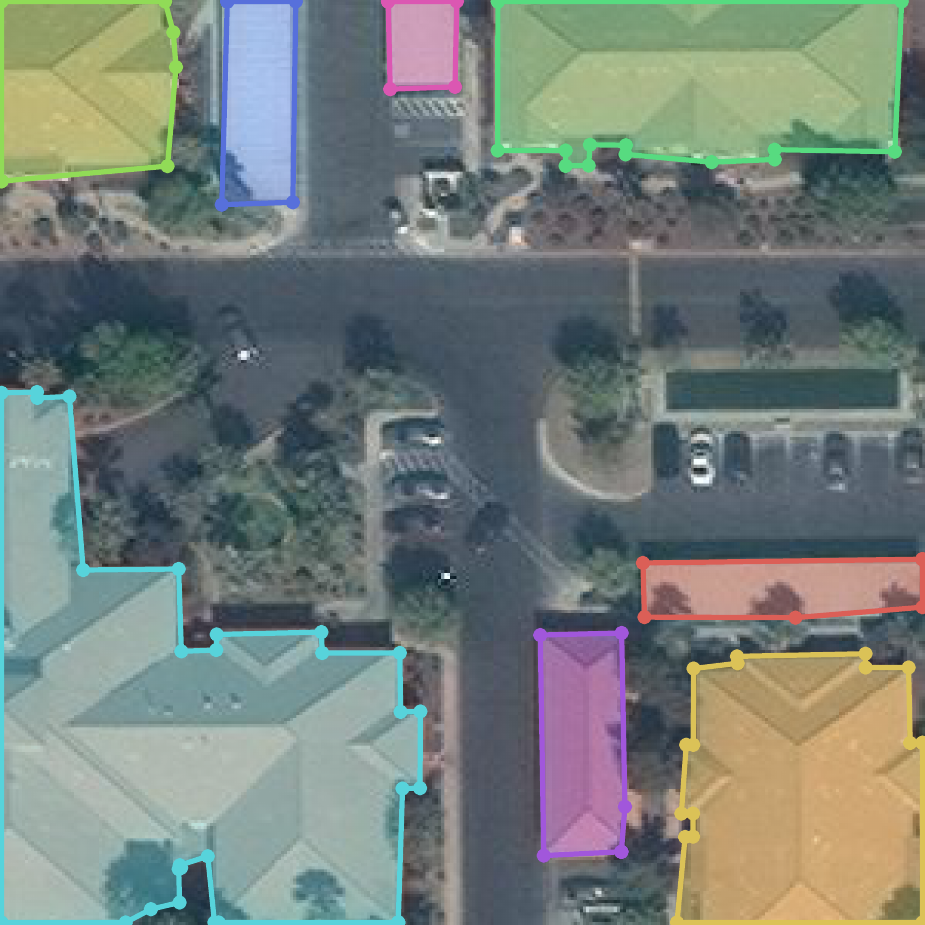}
\label{fig:teaser_fff}}
\subfigure[HiSup Learning (Ours)]{
\includegraphics[width=0.32\linewidth]{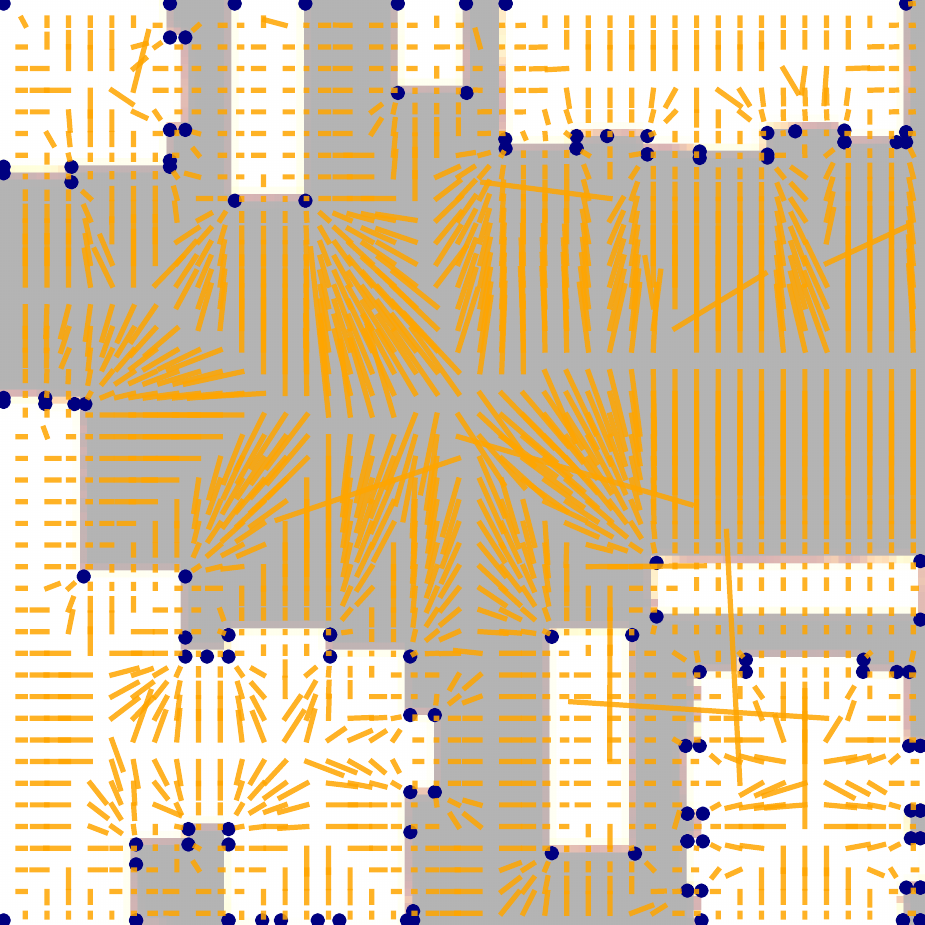}
\includegraphics[width=0.32\linewidth]{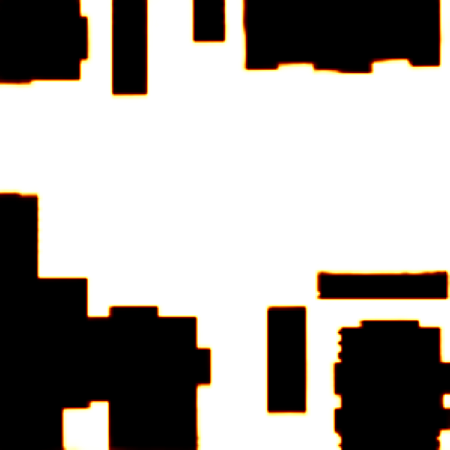} 
\includegraphics[width=0.32\linewidth]{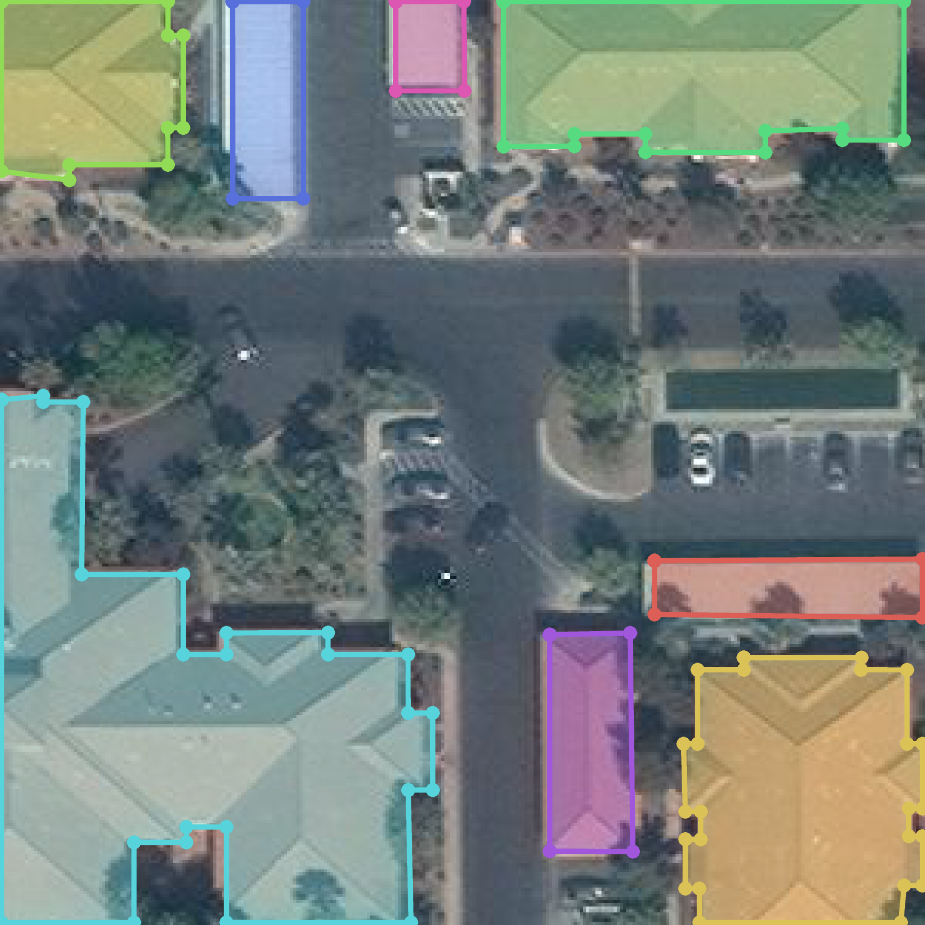}
\label{fig:teaser_ours}}
\caption{
An illustrative comparison between the prior art, Frame-Field Learning~\citep{Girard2020polygonal} and our proposed \emph{HiSup} Learning for polygonal mapping of buildings. In contrast to the Frame-Field that is learned as a complementary to handle the uncertainty of the predicted building masks, our proposed method learns the shape-aware feature embedding by end-to-end learning with \emph{Hierarchical Supervisions}, thus minimized the gap between the predictions of masks (in the middle column) and polygons (in the right column).
}
\vspace{-1cm}
\label{fig:teaser}
\end{center}
\end{figure}

Although the performance of polygonal mapping of buildings has been significantly improved by achieving higher quality segmentation masks, it is worth noticing that the task of polygonal mapping of buildings largely depends on the quality of learned masks. Furthermore, when we get a closer look at the polygonal annotation of buildings, we will find that the learned masks remained an issue of reversibility to the polygonal representation, saying \emph{mask reversibility}. More precisely, with a polygonal annotation of a building, the mask representation is accessible by drawing pixels into an image domain and can be reversed to the polygons as long as the resolution of the image domain is in a reasonable range. However, when we use the mask representation of the polygonal annotations as the supervision signals for the optimization of a convolutional neural network, it would be extremely challenging to obtain such reversible masks.

To handle this \emph{mask-reversibility} challenge, the most recent studies either exploited the shape attributes of buildings by Frame-Field~\citep{Girard2020polygonal} or improved the polygonal mapping of buildings schemes~\citep{Li2020appro} with the off-the-shelf mask predictions, pushed the accuracy of predictions to a higher stage. Despite that, it is also worth noticing that there is a notably large performance gap on the AICrowd benchmark~\citep{Mohanty2020deep} between the polygon and mask predictions. The winning solution~\citep{Jaku2018winner} achieves the mask average precision of 79.1\%, whereas the best-performing model of Frame-Field~\citep{Girard2020polygonal} who was applied to the masks achieves the average precision of 67.0\%. As the mask predicted in Fig.~\ref{fig:teaser_fff} is ambiguous in the shape boundaries, the finally predicted polygons of Frame-Field~\citep{Girard2020polygonal} is still far away from that of human annotations. One possible reason leading to such a large performance gap is that the used supervision signals pay more attention to the high-level regional information rather than the detailed geometric shape of buildings. 

In this paper, we make efforts to close the gap between the polygonal representations and the mask ones of buildings with deep convolutional neural networks, inspired by an observation: 
\begin{quote}
\em 
    The shape composition of polygonal buildings is made up of {\bf points} (or vertices) at the bottom, uses {\bf line segments} to associate the points as the middle-level information, and finally forms the {\bf regional instances} of buildings at high-level semantics.
\end{quote}

Therefore, we are motivated to build the full\hyp{}level supervision to train the neural networks for the sake of shape correctness of the predicted masks. As long as the masks predicted by convolutional neural networks attain the best information about the polygonal shapes, only a simple post-processing scheme is required to achieve the goal of accurate polygonal mapping of buildings.

Towards this goal, we are inspired by the dual representation of line segments, named {\em regional attraction fields}~\citep{xue2019learning,xue2021pami} at the middle-level description as the bridge to get the connection between the unary information of points and complex geometry of regional shapes. Different from the representation of edge pixels, the dual regional representation of line segments demonstrates advantages in two aspects:
\begin{itemize}
    \item It intrinsically connects to the vertices of polygons as they can be viewed as the endpoint of line segments.
    \item The non-local support region of each line segment is corresponding to the region-level information of a part of the building mask.
\end{itemize}

Although the properties of regional attraction fields~\citep{xue2019learning,xue2021pami} are appealing and promising for polygonal mapping of buildings, several challenging open problems remain in computing the higher-ordered geometric shapes from the mid-level line segments (or its dual representation). One may come up with solutions by directly leveraging the detected line segments (or wireframes) as the basic computing clues. However, this will lead to an NP-Hard problem of finding all the closed loops from an undirected graph. Meanwhile, the detection accuracy of line segments and wireframes will also affect the final results. In the most recent, PolyWorld~\citep{zorzi2021polyworld} attempted to solve the problem of polygonal mapping of buildings along this line by using Sinkhorn~\citep{marco2013sink} to approximate the step of finding circulars in a graph. However, it still remained the issue of flexibility when handling the complicated buildings that have holes. 

In the study of this paper, we present a hierarchical supervision (\emph{HiSup}) learning scheme to address the \emph{mask reversibility} issue and show that the geometric information of polygons should be used inductively in the embedding space of the multi-head predictions by end-to-end learning. Specifically, we design a novel information aggregation mechanism to enhance the feature embedding for both the bottom-level vertex prediction and the high-level mask prediction. By explicitly encoding and aggregating the boundary information into the high-level mask head, we successfully reduced the gap between the mask predictions and the expected polygonal ones. During the inference, we design a simple greedy scheme to polygonize and simplify the masks by using the predicted vertices. Our proposed method, \emph{i.e., HiSup}, achieves the state-of-the-art of polygonal mapping of buildings. 

In the experiments, we evaluate our method on the challenging benchmarks of AICrowd~\citep{Mohanty2020deep} and Inria~\citep{maggi2017can}. 
Compared to the current state-of-the-art approaches Frame-Field~\citep{Girard2020polygonal} and the method of \citep{Li2021joint}, the building polygons computed by our \emph{HiSup} obtain a significant improvement of ${\text{AP}}$~\citep{Lin2014microsoft} by 12.4 points and 5.6 points, respectively. In terms of the more challenging metric, $\text{AP}^{\text{boundary}}$~\citep{Cheng2021boundary}, we push the state-of-the-art performance of 50.0\% obtained by PolyWorld~\citep{zorzi2021polyworld} to $66.5\%$. On the Inria dataset~\citep{maggi2017can}, our proposed \emph{HiSup} also obtains a considerable improvement over the Frame-Field~\citep{Girard2020polygonal}.
A systematic ablation study is performed to further justify the proposed method.

The main contributions of this paper are summarized as:
\begin{enumerate}
    \item We address the \emph{mask reversibility} issue of polygonal mapping of buildings by proposing the \emph{HiSup}, which takes the hierarchical supervision signals of vertices, boundaries, and masks to guide the convolutional neural network to learn in a shape-aware fashion.
    \item We present a key component of aggregating the embedding of attraction fields into the learning of high-level masks and the bottom-level vertices. 
    \item We set several new state-of-the-art performances on the challenging AICrowd benchmark in terms of AP and AP$^{\text{boundary}}$ by learning semantically-accurate and geometrically-precise masks with our \emph{HiSup}.
\end{enumerate}

The rest of the paper is organized as follows. Section~\ref{related work} reviews related works. Section~\ref{representation} describes the hierarchical representation. Section~\ref{method} introduces our polygonal mapping of buildings method. Section~\ref{experiment} reports experimental results and some discussions. Finally, conclusions are drawn in section~\ref{conclusion}.

\section{Related work}
\label{related work}
The problem of (polygonal) mapping of buildings is a longstanding problem, which was extensively studied by the means of using hand-crafted features such as textures, geometry, spectrum, and shadows with a vast body of literature (\citet{pesaresi2011improved,Xia2019geosay,ngo2016shape,senaras2013fusion,mannok2015orient}, to cite a few). These kinds of methods could get good results on certain buildings that satisfy the predefined rules, but they suffer from the degraded performance when buildings have inconsistent appearances in complex environments. Therefore, the research focusing on polygonal mapping of buildings has been gradually moved to the deep learning solutions to handle the more challenging real-world data. In this section, we mainly review the recent advancements for learning-based polygonal mapping of buildings. 

\subsection{Learning Building Masks with ConvNets}
The mask representation is widely used for building extraction in two mainstream pipelines: the building map segmentation~\citep{yang2018mapping} and the building instance segmentation~\citep{chen2019darnet}. The former one formulates the problem of building extraction as a pixel-wised binary classification task and the latter one leverages the advances of detected bounding boxes of buildings to directly extract building mask for each instance. Those two pipelines correspond to the semantic segmentation~\citep{long2015fully} and instance segmentation~\citep{He2017maskrcnn} respectively in natural images, whereas, they have the same goal for the task of building extraction in satellite images. As the orthorectified satellite imagery eliminates the occlusions between building instances, it is straightforward to delineate the foreground pixels from the instance-agnostic segmentation map into multiple instances by the connected components labeling schemes~\citep{vladimir2018ternaus,Jaku2018winner}.

In general, learning a ConvNet is promising to obtain high-quality building masks from satellite images~\citep{saito2016multiple,alshehhi2017simultaneous,maggi2017conv,Yuan2018learning}. However, focusing mainly on the high-level semantics makes these approaches perform inaccurate in terms of the building shape correctness. There are some works that focus on alleviating the boundary uncertainties of building maps including multi-scale feature fusion~\citep{maggi2017high,ji2019fully}, and geometrically-aware learning~\citep{li2022building} with the assistance of attraction fields~\citep{xue2019learning,xue2021pami} or edge constraint~\citep{liu2021multiscale}. However, their results are not satisfactory for the accurate polygonal mapping of buildings.

In our method, we concentrate on the geometric correctness of building masks learned by ConvNets. Compared to those methods, our method learns a unified embedding space by exploiting the different level of building representations, which maintains the accuracy of the segmentation masks in both aspects of high-level semantics and the geometric correctness.

\subsection{Polygonal Mapping of Buildings}
Polygonal mapping of buildings refers to getting vectorized building instances, which is the most compact and meaningful representation format beyond others. Only ordered building vertices are kept to describe individual building instances. Traditionally, building vectorization tends to be a separate process after getting the raster map of building instance segmentation. One of the basic algorithms is the Douglas\hyp{}Peucker simplification serial techniques~\citep{douglas1973al}. Those simplification algorithms are usually rough and have difficulty in identifying right building vertices. Extra refine strategies based on the empirical building shape are proposed in~\citep{Wei2020toward}. Lately, based on both the image and its semantic probability map, a polygonal partition approach shows promising results for building vectorization~\citep{Li2020appro}. However, these multi\hyp{}staged methods mostly demand separate optimization processes. An overall optimized result is hard to get.

Currently, automatic polygonal mapping of buildings from remote sensing imagery has received considerable attention. Some methods tend to combine the vectorization process with segmentation through the multi\hyp{}task learning. By learning building segmentation map aligned with proposed frame field vectors, Frame-Field~\citep{Girard2020polygonal} gets building polygons through corner\hyp{}aware contour simplification. The invented frame field map is a 4D vector calculated according to the direction of building edges. In comparison with the frame field~\citep{Girard2020polygonal}, the attraction fields~\citep{xue2019learning,xue2021pami} adopted in our method have more contextual and structural information. \citet{Li2021joint} generate building masks, corner locations, and edge orientations simultaneously. Then the corner location map and edge orientation map are combined together to help with identifying vertices from pixels belonging to building boundaries. Beyond that, an additional refinement process is adopted to further adjust building vertices on a finer scale. The refinement of building vertices can also be embedded to the whole model and be trained jointly~\citep{Chen2020polygoncnn}. 

The above methods all get building polygons through simplifying the predicted building contours. Instead, we use detected vertices to rebuild the whole building polygons. 

There are other methods that extract building polygons on the basis of object detection formula. Buildings are first marked with bounding boxes. Then, for every building, corners are predicted one by one through recurrent neural networks (RNNs) with convolutional long-short term memory (ConvLSTM)~\citep{Li2019topo}. This framework is further improved by~\citep{zhao2021building} with modified modules including global context block (GCB), boundary refinement block (BRB) and gated recurrent units (GRU). Following the same detection formula, \citet{liu2021afmrnn} proposed AFM-RNN which introduced the attraction fields~\citep{xue2019learning} as embedded features to enhance the building corner detection. In our method, we take advantage of the attraction fields as the bridge to connect both the building masks and vertices implicitly. 

Lately, \citet{zorzi2021polyworld} came up with a new proposal termed PolyWorld by considering all building polygons within one image as a whole undirected graph. They predict building polygons by detecting vertices and solving their adjacent connections. The obtained building polygons are very neat in visual. Failed cases appear where buildings have complex structures like inner yards.

\section{A Hierarchical Representation of Buildings}
\label{representation}

\begin{figure*}
    \centering
    \subfigure[Building Polygons]{\begin{minipage}[b]{0.22\textwidth}
			\includegraphics[width=1\textwidth]{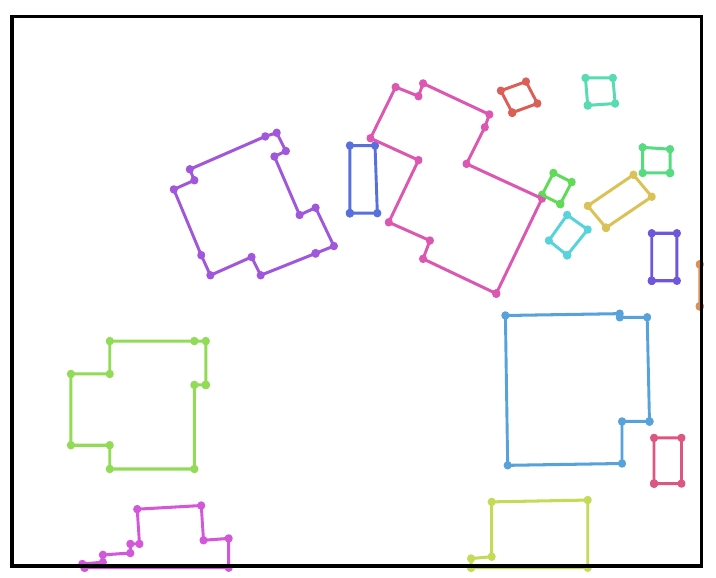}
		\end{minipage}
		\label{fig:represent_poly}}
    \subfigure[Segmentation Map]{\begin{minipage}[b]{0.22\textwidth}
			\includegraphics[width=1\textwidth]{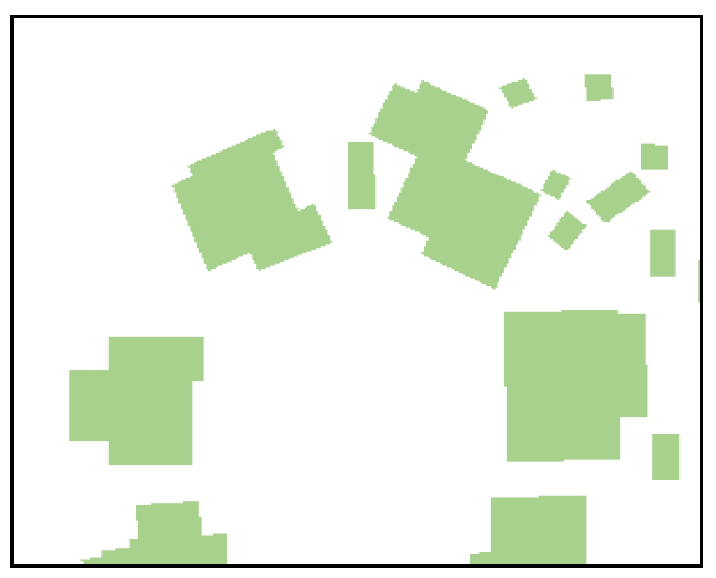}
		\end{minipage}
		\label{fig:represent_seg}}
	\subfigure[Attraction Field Map]{\begin{minipage}[b]{0.22\textwidth}
			\includegraphics[width=1\textwidth]{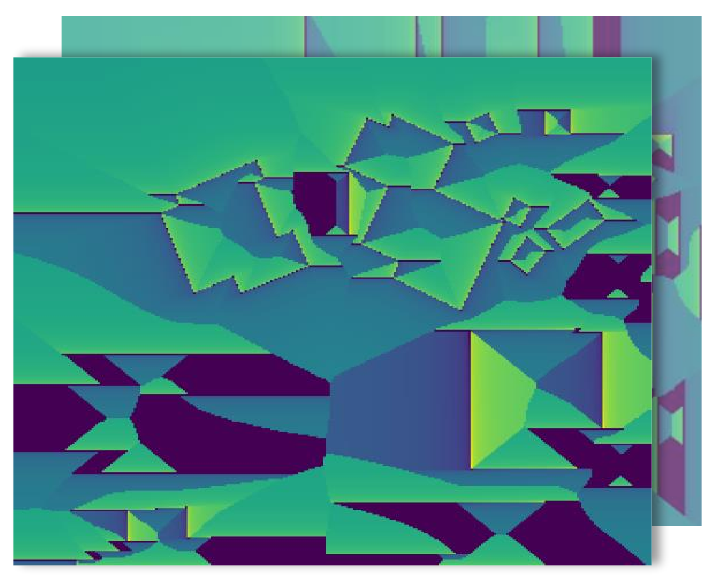}
		\end{minipage}
		\label{fig:represent_afm}}
	\subfigure[Convex/Concave Vertices]{\begin{minipage}[b]{0.22\textwidth}
			\includegraphics[width=1\textwidth]{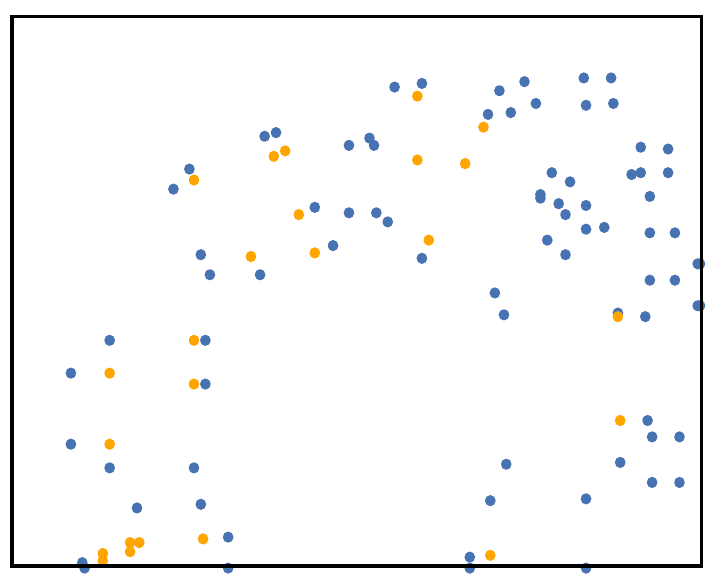}
		\end{minipage}
		\label{fig:represent_junc}}
    \caption{An illustration for the hierarchical representation of vectorized building maps. For the expected building polygons in \subref{fig:represent_poly}, we learn the high-level segmentation masks~\subref{fig:represent_seg}, the mid-level attraction fields of line segments~\subref{fig:represent_afm} and the bottom-level vertices~\subref{fig:represent_junc} with a single convolutional neural network, which guides the neural network learns a unified embedding for accurate polygonal mapping of buildings.
    }
    \label{fig:represent}
\end{figure*}

Given a satellite image $I$ defined on the image lattice $\Lambda = \mathbb{Z}_{W}\times \mathbb{Z}_{H}$ with the resolution of $H\times W$, the $n$ building polygons presented in $I$ are the regions $\Omega_1, \ldots, \Omega_n \subset \mathcal{D} \subset \mathbb{R}^2$, where $\mathcal{D}$ is a continuation of the set $\Lambda$ in the 2D real space. As the buildings are man-made objects, the boundary $\partial \Omega_i$ can be described by a polygon that consists of an array of $m_i$ vertices, denoted by $\partial\Omega_i^{poly} = [\mathbf{x}_i^1,\mathbf{x}_i^2, \ldots, \mathbf{x}_i^{m_i},\mathbf{x}_i^{m_i+1}]$, where $\mathbf{x}_i^{m_i+1} = \mathbf{x}_i^1$. The order of the vertices in $\partial\Omega_i^{poly}$ indicates the adjacent relationship among the vertices.
 
\subsection{High-Level Regional Masks of Buildings}
The mask representation is extensively used for the extraction of buildings as a high-level semantic segmentation task. In this sense, the $n$ regions of building polygons are represented by a segmentation map $\mathcal{S}$ defined on the image lattice $\Lambda$ by
\begin{equation}
    \mathcal{S}(\mathbf{x}) 
    =\begin{cases}
    1, &\text{if } \mathbf{x} \in \Omega_i\,,\\
    0, &\text{otherwise\,.}
    \end{cases}
\end{equation}

Benefiting from the bird-eye view observation of satellites, the $n$ regions can be observed without occlusion between instances, therefore a high-performing convolutional neural network for image semantic segmentation can obtain the instance-level segmentation results in most cases. However, due to the concentration of high-level semantics, the predicted segmentation maps usually pay more attention to the interior regions of $\Omega_i$ rather than their boundaries $\partial\Omega_i$. As a result, it is very challenging to compute accurate polygons from the predicted masks.

\subsection{Bottom-Level and Mid-Level Geometries}
As one main goal of building segmentation is for the vectorized representation, the polygonal representation about the boundary information $\partial\Omega$ is of great importance for the end task. Here, we briefly discuss both the bottom-level vertices and the mid-level line segments about the boundary information.

When we focus on the polygonal boundary of the buildings, the polygon $\partial\Omega_i^{poly}$ for the $i$-th building mask contains two kinds of information: the bottom-level vertices $\{\mathbf{x}_i^1,\ldots, \mathbf{x}_i^n\}$ and their adjacent relationship represented by mid-level line segments \(\{\mathbf{l}^1_i,\ldots,\mathbf{l}^n_i\}\). 
With the bottom-level and mid-level geometries of buildings, there would be another challenging problem of grouping primitives into instances. It was recently approached under the view of graph computation with Sinkhorn approximation~\citep{marco2013sink} to avoid the NP-Hard problem of finding circulars in an undirected graph~\citep{zorzi2021polyworld}.  

In this paper, we believe that although the mask learning is a high-level recognition task, as the polygonal buildings can be lossily converted into mask representation, there should be a unified embedding space to learn and decode consistent multi-level outputs.
To approach this, we use the bottom-level vertices and mid-level line segments as the complementary information with the learning of masks, which guides the convolutional neural networks to focus on the bottom-level and mid-level information for the shape correctness of the masks.

\paragraph*{Convex and Concave Building Vertices.}
Inspired by \citet{Li2021joint}, we divide the building vertices (~\emph{i.e.}, junctions) according to their convexity. In detail, for the vertices in $\partial\Omega_i^{poly}$, vertices belonging to the minimal convex hull of $\partial\Omega_i^{poly}$ are regarded as the convex vertices, denoted in $V_{\text{convex}}^i$. For the vertices that belong to the set $\partial\Omega_i^{poly}/V_{\text{convex}}^i$, they are denoted by $V_{\text{concave}}^i$. By stacking all the vertices across instances, we have two sets $V_{\text{convex}}$ and $V_{\text{concave}}$ as the bottom-level representation of buildings.

\paragraph*{Regional Encoding of Line Segments.}
As discussed before that the line segments are actually the bridge between the bottom-level vertices and the high-level masks, we are interested in the representation of line segments that are well-suited for polygonal mapping of buildings. The most natural choice of the line segments might be the edge map (or line heatmap). However, we found that the learning of edge maps can only improve the mask reversibility marginally. Therefore, we are motivated to use the dual representation, the Attraction Field Representation Map (AFM)~\citep{xue2019learning,xue2021pami} for our endtask. Denoted by $L = \{\mathbf{l}_i=(\mathbf{x}_i^s,\mathbf{x}_i^e)\}_{i=1}^m$ (\(\mathbf{x}_i^s\) and \(\mathbf{x}_i^e\) are the two endpoints), the line segments of the polygonal instances $\{\partial\Omega_i\}$ partition the image lattice $\Lambda$ into $m$ regions $\mathcal{R} = \{R_i\}_{i=1}^m$ by,
\begin{equation}
    R_i = \{\mathbf{x}\in \Lambda|d(\mathbf{x},\mathbf{l}_i)<d(\mathbf{x},\mathbf{l}_j),\forall j\neq i,\mathbf{l}_j\in L\}\,,
\end{equation}
\begin{equation}
    d(\mathbf{x}, \mathbf{l}_i) = \min_{t\in[0,1]}\|\mathbf{x}_i^s+t\cdot(\mathbf{x}_i^e-\mathbf{x}_i^s)-\mathbf{x}||_2^2\,.
\end{equation}

With the region partition $\mathcal{R}$, the attraction field map $\mathcal{A}\in\mathbb{R}^{2\times H\times W}$ between any pixel to its close-in line segment $\mathbf{l}_j$ is computed by 
\begin{equation}
    \mathcal{A}(\mathbf{x}) = \mathbf{x}' - \mathbf{x}\,,
\end{equation}
where the point $\mathbf{x}'$ is the projection of $\mathbf{x}$ on $\mathbf{l}_j$. As the attraction field map encodes the line segments according to $\mathcal{R}$, the high-level masks are implicitly encoded.

\section{Learning Reversible Building Masks}
\label{method}
In this section, we present a novel approach that learns building masks in a unified embedding space under the supervision of masks, line segments and the vertices. The unified embedding space takes the merits of different-level of information for the buildings, thereafter enables the convolutional neural networks to learn both the instance-level information and the geometric details with the best reversibility for polygonal mapping of buildings. Then, we present a mask attraction scheme that polygonizes the reversible masks into the expected polygons by attracting the sparse vertices to the boundary of masks.

\begin{figure}
\begin{center}
\includegraphics[width=0.5\textwidth]{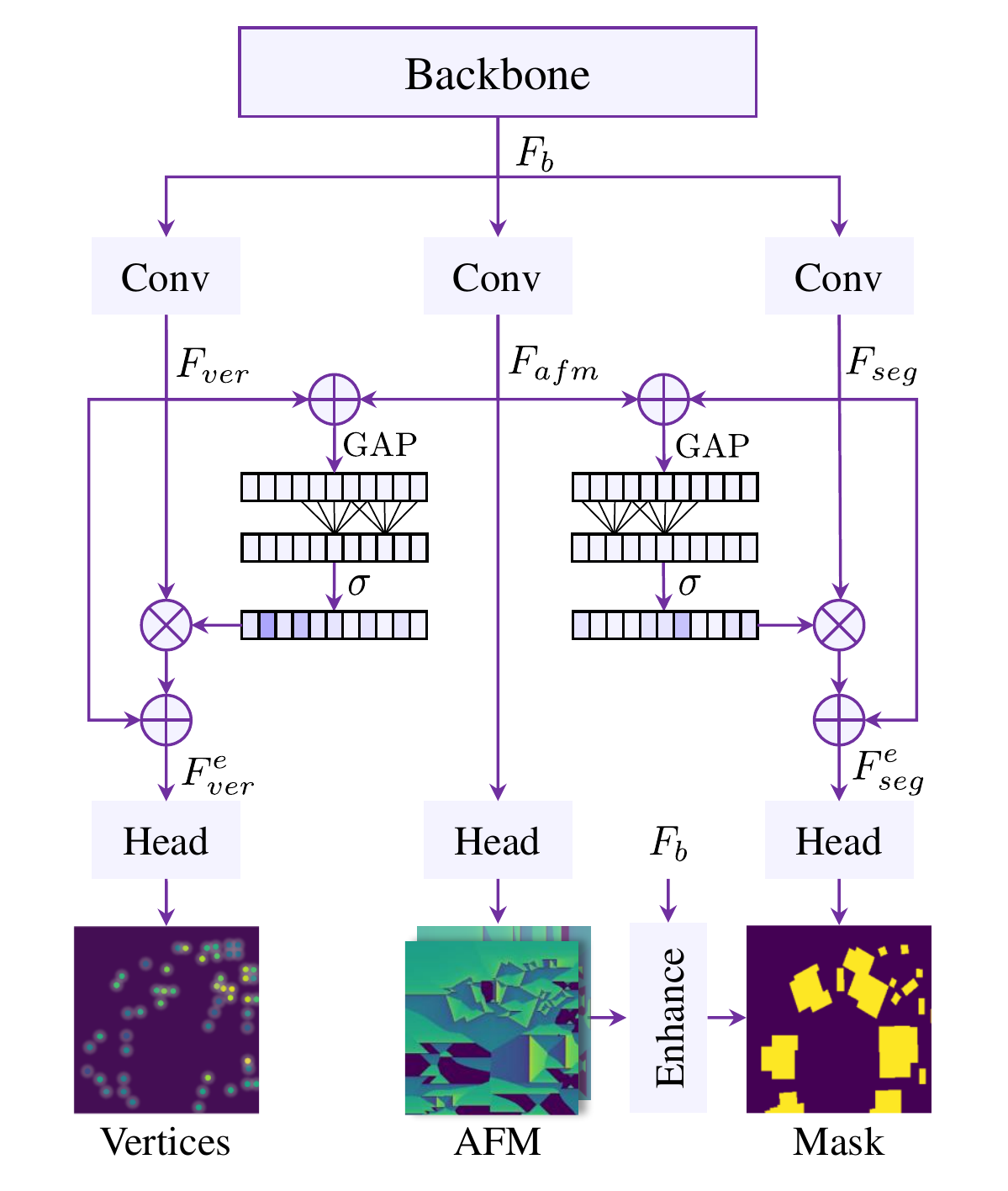}
\caption{The overall training process. The ``Conv'' module denotes three continuous sequences including \(3\times 3\) convolution, batch normalization and ReLU layers. The ``Head'' module denotes two \(3\times 3\) and \(1\times 1\) convolution layers connected with an activation layer. $\otimes$ denotes element-wise product and $\oplus$ denotes element-wise addiction. The ``Enhance'' module denotes the concatenated operations mentioned in Geometric-Aware Mask Learning.}
\label{fig:train}
\end{center}
\end{figure}

\subsection{Network Architecture Overview}
As illustrated in Fig.~\ref{fig:train}, given an image $I \in \mathbb{R}^{3\times H\times W}$, we use a backbone network (\emph{e.g.,} HRNets~\citep{Wang2021hrnet} or UResNets (Unet~\citep{ronneberger2015u} with ResNet~\citep{he2016deep} encoder)) to extract the shared feature map $\mathbf{F}_b \in \mathbb{R}^{3\times H_s\times W_s}$, where $H_s = \lfloor H/s \rfloor$ and $W_s = \lfloor W/s \rfloor$ are the size of $\mathbf{F}_b$ with the down-sampling factor $s$. Because our method is learned from the bottom-level to high-level information of buildings, three convolution layers (including the BatchNorm~\citep{ioffe2015batch} and ReLU~\citep{nair2010rectified}) with 256 output channels are used to transform the backbone feature $\mathbf{F}_b$ into the embedding features $\mathbf{F}_{\text{ver}}$ of vertices, $\mathbf{F}_{\text{afm}}$ of the attraction field maps, and $\mathbf{F}_{\text{seg}}$ of segmentation maps. The different embedding features are then transformed by the cross-level interaction module (introduced in Sec.~\ref{sec:CLIL}) to predict (1) the heatmap $\mathbf{H}$ and the short-range offset field $\mathbf{O}$ for the vertices, (2) the attraction field map $\mathbf{A}$ of the line segments, and (3) the segmentation mask $\mathbf{S}$ for the segmentation map of the building polygons.

\subsection{Learning from Cross-Level Interactions}\label{sec:CLIL}
As the attraction field representation~\citep{xue2019learning,xue2021pami} has strong correlations with the vertices and segmentation masks, we are going to regularize the feature maps $\mathbf{F}_{\text{seg}}$ and $\mathbf{F}_{\text{ver}}$ to emphasize their shape correctness for the mask predictions and the semantic consistency in the learning of vertices. 
\paragraph*{Efficient Channel Attention with AFM Embedding.}
Inspired by the success of ECA-Net~\citep{wang2020eca}, we extend the efficient channel attention mechanism to enhance the feature representation of $\mathbf{F}_{\text{seg}}$ and $\mathbf{F}_{\text{ver}}$ by 
\begin{equation}
    \begin{split}
        \mathbf{F}_{\text{seg}}^e &= \sigma(\text{C1D}(\text{GAP}(\mathbf{F}_{\text{afm}} + \mathbf{F}_{\text{seg}})))\cdot \mathbf{F}_{\text{seg}}
        + \mathbf{F}_{\text{seg}}\,,\\
        \mathbf{F}_{\text{ver}}^e &= \sigma(\text{C1D}(\text{GAP}(\mathbf{F}_{\text{afm}} + \mathbf{F}_{\text{ver}})))\cdot \mathbf{F}_{\text{ver}}
        + \mathbf{F}_{\text{ver}}\,,
    \end{split}
\end{equation}
where $\text{C1D}(\cdot)$, $\text{GAP}(\cdot)$, $\sigma(\cdot)$ are the 1-D convolution, global average pooling and Sigmoid function.

\paragraph*{Geometric-Aware Mask Learning.}
We lift the learned attraction field maps by a linear layer from the predicted attraction field $\mathbf{A}\in\mathbb{R}^{2\times H_s\times W_s}$ to a 128-channel tensor $\mathbf{F}_{\text{afm}}^*$ and concatenate it to the backbone feature $\mathbf{F}_b$. The concatenated feature map is subsequently used to predict the building masks $\mathbf{S}_{\text{afm}}$, which guides the backbone feature $\mathbf{F}_b$ to learn the shape correctness from the target of attraction field maps via backpropagation. As the backbone feature map is regularized by the predicted attraction field $\mathbf{A}$, we are able to use the enhanced feature map $\mathbf{F}_{\text{seg}}^e$ for the final mask prediction, denoted by $\mathbf{S}_f$. During training, the predicted intermediate mask $\mathbf{S}_{\text{afm}}$ and the final mask $\mathbf{S}_{f}$ are supervised by the binary cross entropy (BCE) loss with the groundtruth masks $\mathbf{S}^*$ by 
\begin{equation}
    \mathcal{L}_{\text{mask}} = \text{BCE}(\mathbf{S}_{\text{afm}}, \mathbf{S}^*) + \text{BCE}(\mathbf{S}_{\text{f}}, \mathbf{S}^*)\,,
\end{equation}
while the attraction field $\mathbf{A}$ is learned from the $\ell_1$ loss function by
\begin{equation}
    \mathcal{L}_{\text{afm}} = \ell_1(\mathbf{A},\mathbf{A}^*)\,,
\end{equation}
where $\mathbf{A}^*$ is the ground truth of attraction fields generated by the edge of polygons.

\paragraph*{Semantic-Consistent Vertices Prediction.}
Instead of using the salient edge pixels (from predicted masks) as building vertices, we design a vertex branch to learn the heatmap and short-range offsets of vertices. Taking the enhanced feature $\mathbf{F}_{\text{ver}}^e$, a convolution layer is used to learn the heatmap $\mathbf{H}$ and the offset map $\mathbf{O}$ by the loss functions of cross entropy and $\ell_1$ loss by 
\begin{equation}
\begin{split}
    \mathcal{L}_{\text{vcls}} &= \text{BCE}(\mathbf{H},\mathbf{H}^*)\,, \\
    \mathcal{L}_{\text{voff}} &= \ell_1(\mathbf{O}\cdot\mathbf{H}^*,\mathbf{O}^*\cdot\mathbf{H}^*)\,, \\
\end{split}
\end{equation}
where $\mathbf{H}^*$ is the ground truth of vertices with the values $0$ for background pixels and $1$ for the vertices targets, $\mathbf{O}^*$ is the short-range offsets that are in the range of $[-0.5,0.5)$ for the target pixels.

\paragraph{Total Loss.}
To train the neural network, we use the predefined loss factors to balance the different types of predictions in the total loss function by
\begin{equation}
    \mathcal{L}_{\text{total}} = \lambda_m\cdot\mathcal{L}_{\text{mask}} + \lambda_a\cdot\mathcal{L}_{\text{afm}} + \lambda_c\cdot\mathcal{L}_{\text{vcls}} + \lambda_o\cdot\mathcal{L}_{\text{voff}}\,,
\end{equation}
where $\lambda_m$, $\lambda_a$, $\lambda_c$ and $\lambda_o$ are the loss factors of the predicted masks, attraction field maps, the heatmaps and offsets of the vertices. In our experiment, we empirically set $\lambda_m = 1.0$, $\lambda_a = 0.1$, $\lambda_c = 8.0$ and $\lambda_o = 0.25$ according to their loss magnitudes.

\begin{figure*}[!b]
\begin{center}
\includegraphics[width=0.9\textwidth]{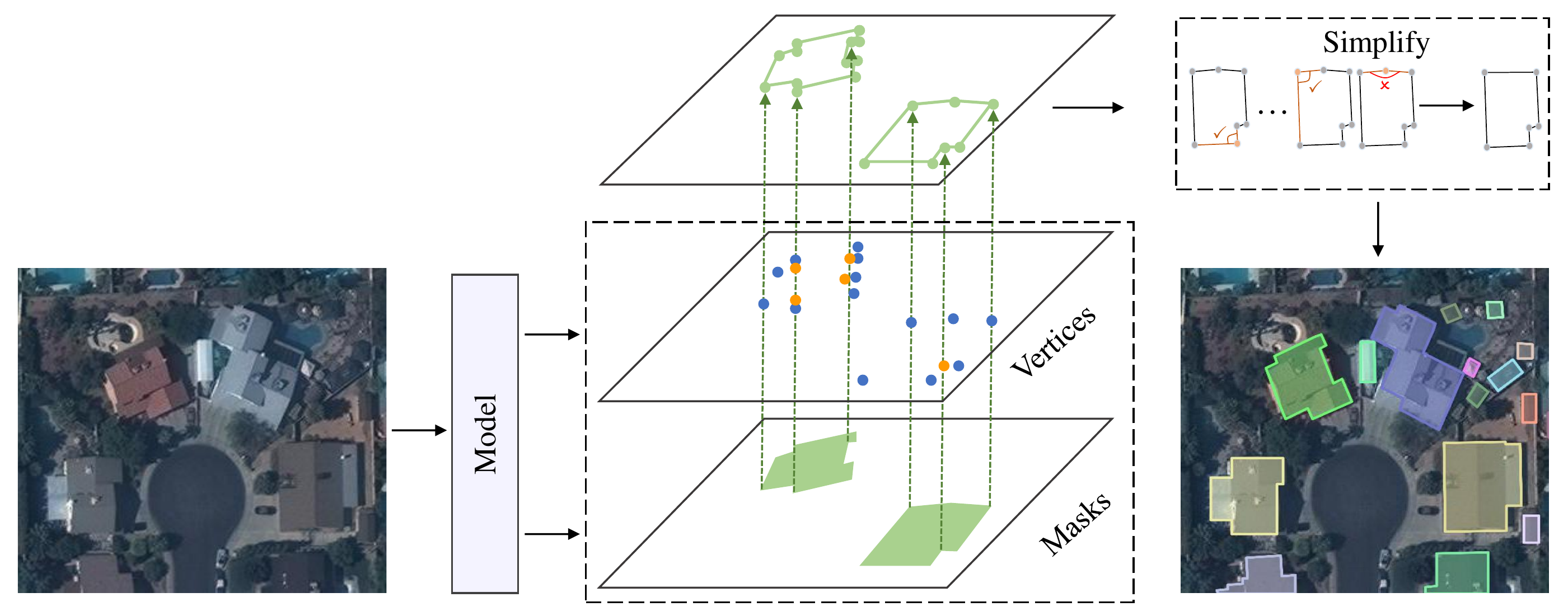}
\caption{The inference process of our method. Building polygons are generated through matching masks with their nearest junctions. The abundant junctions that do not belong to building vertices are removed afterwards.}
\label{fig:inference}
\end{center}
\end{figure*}
\subsection{Extracting Polygons by Mask Attraction}
Benefiting from our design of cross\hyp{}level interaction 
module, the learned masks have a desired property of shape correctness in the boundary geometry. Therefore, the polygonization of the building masks would be straightforward by tracing the boundary pixels from the predicted masks. For the sake of computing the most simplified polygons without redundant vertices, we found that the learned vertices from $\mathbf{H}$ and $\mathbf{O}$ cooperate very well by attracting the traced boundary pixels of masks to the instance-agnostic vertices. We term our scheme of polygonization as Mask-and-Vertices Attraction, short in \emph{MaV-Attr}. The whole pipeline of our polygon extraction is shown in Fig.~\ref{fig:inference}.

\paragraph*{Polygon Initialization.} Given a predicted mask $\mathbf{S}$, the corresponding heatmap $\mathbf{H}$ and offset $\mathbf{S}$ of vertices, we extract the building instances by a given threshold of $\tau \in (0,1)$ to get the binary mask prediction $\mathbf{S}_{\tau} = \mathbf{S}>\tau$, which are subsequently processed into $n$ instances by computing the connected components, denoted by $\Omega_1, \ldots, \Omega_n$. The redundant polygonal representation of $\Omega_i$ is computed by tracing all the boundary pixels, denoted in $\partial\Omega_i^{init} = [\mathbf{x}_1^i,\ldots,\mathbf{x}_r^i, \mathbf{x}_{r+1}^i = \mathbf{x}_1^i]$, where $r$ is the number of boundary pixels of $\Omega_i$. With the initial redundant polygons $\partial\Omega_i^{init}$, we only need to focus on its simplification by sparse vertices that are computed by leveraging the local non-maximum suppression (NMS) in $3\times 3$ neighborhoods. The local NMS is implemented by a MaxPooling layer with the kernel size $3\times 3$.
After the local NMS, the top-$K$ ($K = 300$ in our experiments) vertices that have a higher classification score than $\tau_v=0.008$ are retained and refined by the offset vectors for sub-pixel accuracy, denoted by $V = \{\mathbf{v}_1,\ldots, \mathbf{v}_m\}$. 

\paragraph*{Polygon Simplification.} 
For the $i$-th initial polygon $\partial_i^{init}$, we match each vertex $\mathbf{x}_k^i$ to the closest vertex $\mathbf{v}_{\pi(k,i)} \in V$ via computing the Euclidean distance by
\begin{equation}
    \pi(k,i) = \arg\limits_j\min_{\mathbf{v}_j\in V} \left\|\mathbf{x}_k^i - \mathbf{v}_j\right\|_2,
\end{equation}
where $\pi(k,i)$ is an index mapping from the $k$-th boundary pixel of the $i$-th initial polygon to the closest vertex in $V$. Then, we go through all the boundary pixels $\mathbf{x}_k^i$ and remove the non-minimal pixels that have the same indices. That is to say, for any boundary pixel $\mathbf{x}_k^u$ and $\mathbf{x}_k^v$, if their indices satisfying $\pi(u,i) = \pi(v,i) = z$ and $\|\mathbf{x}_k^u - \mathbf{v}_z\| < \|\mathbf{x}_k^v - \mathbf{v}_z\|$, the pixel $\mathbf{x}_k^v$ is removed from the initial polygon $\partial_i^{init}$. 
In addition to this, for the pixels that are far away from any predicted vertex (under a given threshold $\tau_d = 5$) are also removed.
As we already know the topological relationship between any boundary pixels in $\partial_i^{init}$, a linear scanning is required to efficiently remove the redundant boundary pixels. In the final step, we check the adjacent edges of polygons and merge them if they are parallel up to an angle tolerance of $10^\circ$.

The simplified polygon of $\partial_i^{init}$ is denoted by $\partial_i^{final} = [\hat{\mathbf{x}}_1,\ldots,\hat{\mathbf{x}}_M,\hat{\mathbf{x}}_{M+1}=\hat{\mathbf{x}}_1]$. For any vertex $\hat{\mathbf{x}_i} \in \partial_i^{final}$, it must be in the set $V$. The number of vertices $M = |\partial_i^{final}|$ is far less than $m = |\partial_i^{init}|$.

\subsection{Implementation details}
In our implementation, we mainly use the HRNetV2-W48~\citep{Wang2021hrnet} as the backbone for our method. Meanwhile, we also use the smaller backbones, HRNetV2-W18, HRNetV2-W32~\citep{Wang2021hrnet}, and the  UResNet101~\citep{ronneberger2015u,he2016deep} to show that the possible usage of our method in more practical configurations. The input images are resized to \(512\times 512\) during both training and testing. As all the used backbone networks will yield the down-sampled feature maps with the down-sampling factor of $s=4$, the output maps of our method are with the resolution of $128\times 128$. In the inference phase, we will resize the polygonal outputs into original image size (\emph{e.g.,} $300\times 300$ in AICrowd dataset~\citep{Mohanty2020deep}) for evaluation. 

During training, we used the data augmentation strategies (as used in~\citep{Girard2020polygonal}) including random flip, random rotation, and color jittering. The ADAM optimizer~\citep{diederik2015adam} is used to train our network on 4 Nvidia RTX 3090 GPUs with the initial learning rate of $10^{-4}$ and the weight decay $10^{-4}$ on PyTorch 1.8. The number of training epochs is set to $100$ and the learning rate is decayed by $10$ after 25 epochs of training. For the ablation study, we reduce the training epochs from $100$ to $30$ due to the limited computation resources and time.

\section{Experiment}
\label{experiment}
\subsection{Datasets and evaluation metrics}
We compared our method to the state\hyp{}of\hyp{}the\hyp{}art methods on two publicly available datasets.

\subsubsection{AICrowd dataset}
AICrowd Mapping Challenge dataset~\citep{Mohanty2020deep} (AICrowd dataset) contains $300\times 300$ pixels RGB images and corresponding annotations in MS-COCO~\citep{Lin2014microsoft} format. The training set includes 280741 tiles. For this challenge is already closed, we use the validation subset as testing, whose amount of tiles is 60317. AICrowd dataset also provided users with a small version that contains smaller amount of samples. We use the small version of AICrowd dataset in the ablation studies. 

For evaluating the performance of building instance segmentation, we follow the standard MS\hyp{}COCO~\citep{Lin2014microsoft} evaluation metrics. They are also the official metrics for AICrowd dataset. Particularly, we report the average precision
(AP) and average recall (AR) metrics under a range of intersection over union (IoU) thresholds from 0.50 to 0.95 with a step of 0.05. The precision and recall scores under IoU thresholds of 50 and 75 are also listed in Table~\ref{tbl_crowdai} denoted as \(\rm{AP}_{50}\), \(\rm{AP}_{75}\), \(\rm{AR}_{50}\) and \(\rm{AR}_{75}\). Moreover, to put extra emphasis on the quality of building contours, we introduce the metrics of Boundary IoU~\citep{Cheng2021boundary}. Compared with MS\hyp{}COCO metrics, the Boundary IoU metrics are more reasonable to reveal the accuracy of the detected instance boundaries. For two masks \(A\) and \(B\), the Boundary IoU only computes the IoU between sets of pixels that are within distance \(d\) from two mask contours. Denoted the contours of masks \(A\) and \(B\) as \(A_d\) and \(B_d\), the Boundary IoU is defined as
\begin{equation}
{\rm Boundary\ IoU} (A,B)=\frac{|(A_d\cap A)\cap(B_d\cap B)|}{|(A_d\cap A)\cup(B_d\cap B)|}\,.
\end{equation}
Here the \(d\) equals 0.02 which is consistent with the setting of COCO instance segmentation. The average boundary precision denoted by \(\rm{AP}^{boundary}\) under a range of IoU to object boundary is reported in Table~\ref{tbl_crowdai_poly}.

All the above evaluation metrics are based on results at the segmentation level. To directly compare the quality of predicted building polygons, we report the evaluation of PoLiS metric~\citep{Avbelj2015metric}.   
For two given polygons A and B, the PoLiS is defined as the average distances between each vertex \(a_j\in A, j=1,\ldots,q\) of A and its closest point \(b\) within the boundary \(\partial B\) and vice versa. Suppose the polygon B has vertices \(b_k\in B, k=1,\ldots,r\), the PoLiS metric can be expressed as
\begin{equation}
\begin{aligned}
{\rm PoLiS}(A, B) =& \frac{1}{2q}\sum_{a_j\in A}\min_{b\in \partial B}||a_j-b||\\
    & +\frac{1}{2r}\sum_{b_k\in B}\min_{a\in \partial A}||b_k-a||\,,
\end{aligned}
\end{equation}
where $\frac{1}{2q}$ and $\frac{1}{2r}$ are the normalization factors. The IoU threshold for filtering predicted building polygons is set to 0.5 as referred to \cite{zhao2021building}.

The IoU and complexity aware IoU (C\hyp{}IoU)~\citep{zorzi2021polyworld} are also calculated for evaluation. For the two polygons \(A\) and \(B\), the C\hyp{}IoU metric is computed as:
\begin{equation}
\text{C-IoU}(A,B)={\rm IoU}(A_m,B_m)\cdot (1-{\rm RD}(N_A, N_B))\,,
\end{equation}
where the first term IoU\((\cdot)\) indicates the normal IoU between two compared polygon masks \(A_m\) and \(B_m\). The second term \({\rm RD}(N_A, N_B))=|N_A-N_B|/(N_A+N_B)\) is the relative difference between the total number of vertices \(N_A\) from polygon \(A\) and the number of vertices \(N_B\) from polygon \(B\). The C\hyp{}IoU metric considers both segmentation accuracy and polygonization complexity.

\subsubsection{Inria dataset}
Inria Aerial Image Labeling dataset~\citep{maggi2017can} (Inria dataset) contains aerial orthorectified color imagery of $5000\times 5000$ pixels with a spatial resolution of 0.3 m. Unlike AICrowd dataset, the training and testing images of Inria dataset come from different cities where the building distribution varies. The original Inria dataset provides annotations of pixel-wise semantic segmentation, which is not suitable for getting polygonized results by our method. During the training process, we use the traditional method to convert the raster ground truth labels to vector format. As suggested in \citep{maggi2017can}, the first five images of each location from the Inria training set are used for validation.  

As the Inria dataset~\citep{maggi2017can} did not public the testing samples, we follow their official evaluation protocol by submitting the polygonized results in the format of segmentation masks to their evaluation server. The recommended metrics of IoU and accuracy (Acc) are adopted for comparison. The Acc for Inria dataset is defined as the percentage of correctly classified pixels, while the IoU is computed only for pixels belonging to buildings.

\subsection{Results and Analysis}
\subsubsection{Results on AICrowd dataset}
Table~\ref{tbl_crowdai} and Table~\ref{tbl_crowdai_poly} demonstrate the quantitative evaluation results on the AICrowd dataset. We compare our method with other competing approaches including Mask R\hyp{}CNN~\citep{He2017maskrcnn} based on the implementation of \citep{Mohanty2020deep}, PolyMapper~\citep{Li2019topo}, ASIP~\citep{Li2020appro}, Frame\hyp{}Field~\citep{Girard2020polygonal}, PolyWorld~\citep{zorzi2021polyworld} and the recent work proposed by \citep{Li2021joint}. The Mask R\hyp{}CNN implemented by~\cite{Mohanty2020deep} serves as the baseline for AICrowd dataset, which can only obtain building masks instead of polygons. To be able to compare with others, a post\hyp{}processing with Douglas\hyp{}Peucker (DP) simplification~\citep{douglas1973al} is adopted to generate vectorized results from the pixel-wise segmentation. The rest of methods all have outputs of polygonal building instances. 

It can be seen in Table~\ref{tbl_crowdai} that our method gets either the highest or comparable scores under the MS\hyp{}COCO metrics. The average precision termed AP reaches 79.4\% by our method. Only the recall score under IoU equals 0.5 is slightly lower than the ASIP~\citep{Li2020appro} method. It is probably due to the omission of some tiny building pieces that are near the edge of images. 

\begin{table*}
\centering
\caption{Comparison of evaluation results of MS\hyp{}COCO metrics on the AICrowd dataset. The highest scores are in bold.}\label{tbl_crowdai}
\begin{tabular}{|l|l|cccccc|}
\toprule
Method & Backbone &AP\(\uparrow\) &$\rm{AP}_{50}\uparrow $ &$\rm{AP}_{75}\uparrow$ &AR\(\uparrow\) &$\rm{AR}_{50}\uparrow$ &$\rm{AR}_{75}\uparrow$ \\
\hline
Mask R-CNN~\citep{He2017maskrcnn,crowdAI2018baseline} &ResNet101 &41.9 &67.5 &48.8 &47.6 &70.8 &55.5 \\
PolyMapper\footnotemark[1]~\citep{Li2019topo} &VGG16 &50.8 &81.7 &58.6 &58.5 &85.4 &67.0 \\
ASIP~\citep{Li2020appro} &- &65.8 &87.6 &73.4 & 78.7 &\textbf{94.3} & 86.1\\
Frame-Field\footnotemark[2]~\citep{Girard2020polygonal} &UResNet101 &67.0 & 92.1 &75.6 &73.2 &93.5 &81.1\\
\citet{Li2021joint} &UResNet101 &73.8 &92.0 &81.9 &72.6 &90.5 &80.7 \\
PolyWorld~\citep{zorzi2021polyworld} &R2U-Net &63.3 &88.6 &70.5 &75.4 &93.5 &83.1\\
\hline
\textbf{HiSup (ours)} & HRNetV2-W48 &\textbf{79.4} & \textbf{92.7} & \textbf{85.3} & \textbf{81.5} & 93.1 & \textbf{86.7} \\
\bottomrule
\end{tabular}
\end{table*}
\footnotetext[1]{The implementation code from the author's github repository (https://github.com/lizuoyue/ETH-Thesis/tree/master/building) is used for training by ourselves. And we perform the testing on the defined testing dataset.}
\footnotetext[2]{Here the results are reported by applying the polygonization method mentioned in Frame-Field~\citep{Girard2020polygonal} to the the segmentation probability maps from the UNet-Variant~\citep{Jaku2018winner}.}

Table~\ref{tbl_crowdai_poly} reports the additional evaluation results of methods provided with code. The Boundary IoU is a more restricted measurement of the building boundary area. The performance of our method ranks first place on the average boundary precision. It indicates that the building polygons obtained by our method have more precise shapes. Apart from the evaluation of instance segmentation, our method also gets 0.726 on the PoLiS metric, which outperforms the other three vectorization methods. It demonstrates that our method has less overall dissimilarity between vertices of predicted building polygons and the ground truth. For the C\hyp{}IoU metric, our method surpasses all the other listed methods, which means that the buildings extracted by our method are represented in both concise and accurate polygons. Meanwhile, the learned reversible building masks give our method more flexibility for representing buildings in subtle structure details.

\begin{figure*}
\centering
\begin{tabular}{rcccc}
\raisebox{50pt}[5pt]{\rotatebox[origin=c]{90}{PolyMapper}}
\includegraphics[height=110pt,width=110pt]{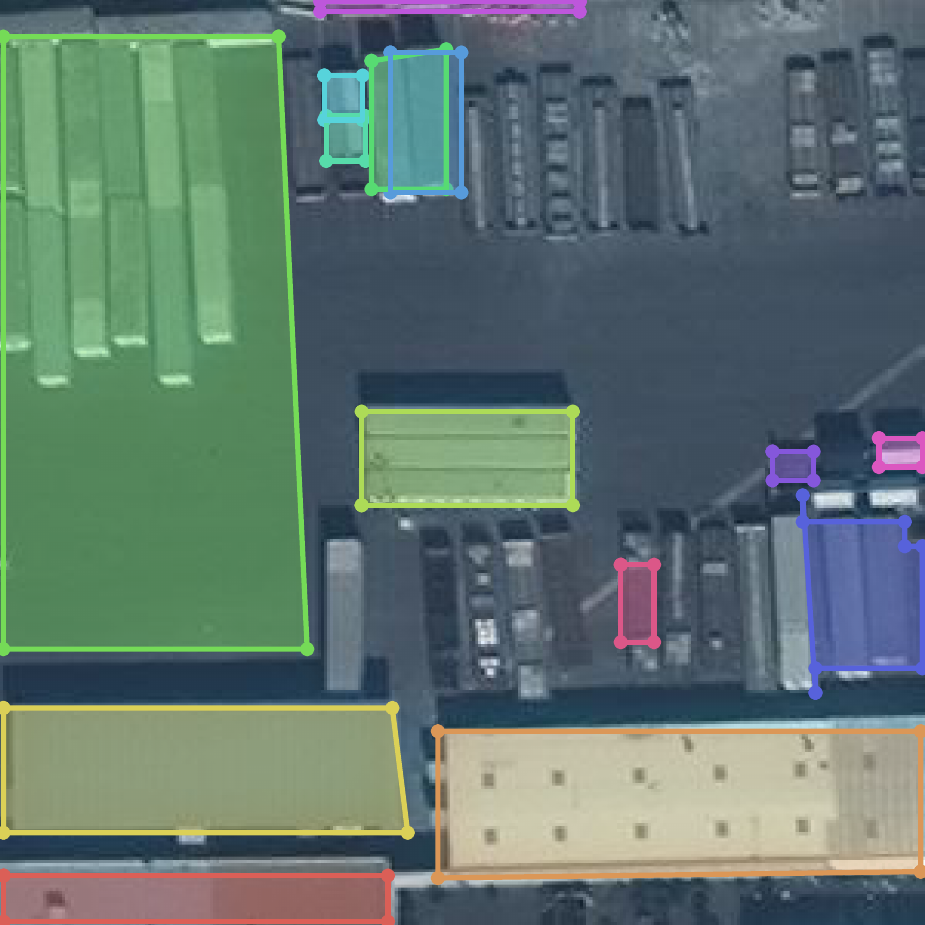}
\includegraphics[height=110pt,width=110pt]{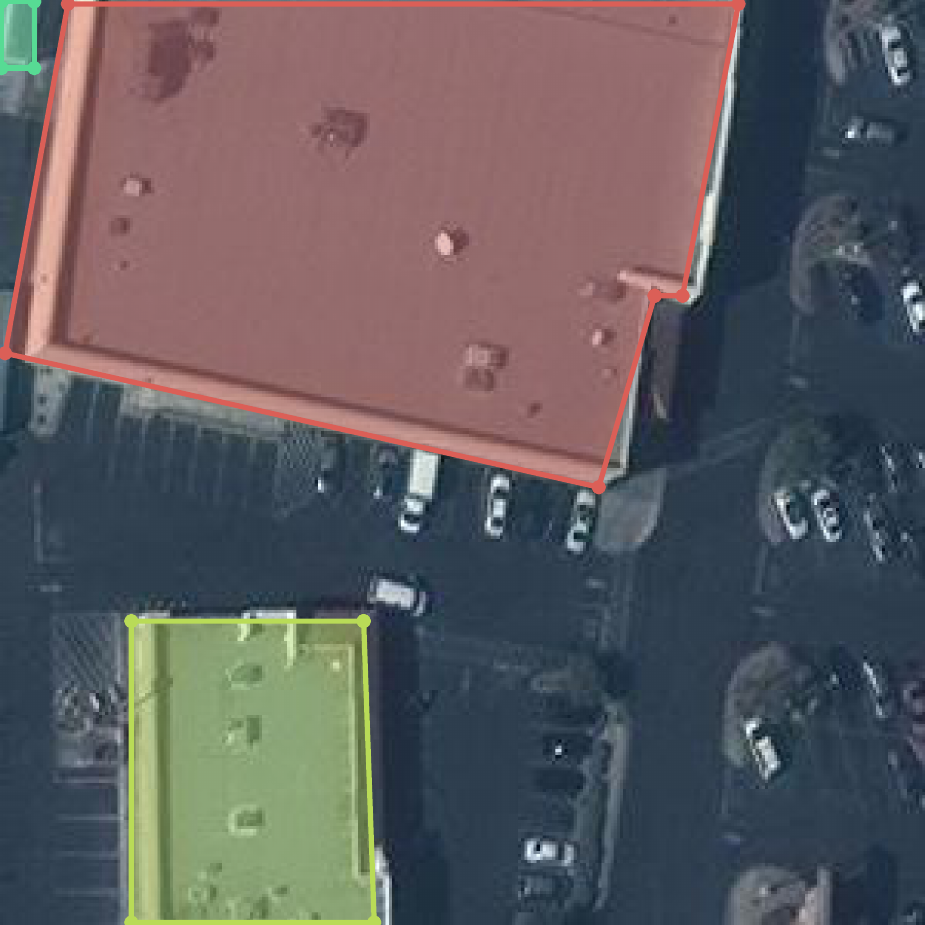}
\includegraphics[height=110pt,width=110pt]{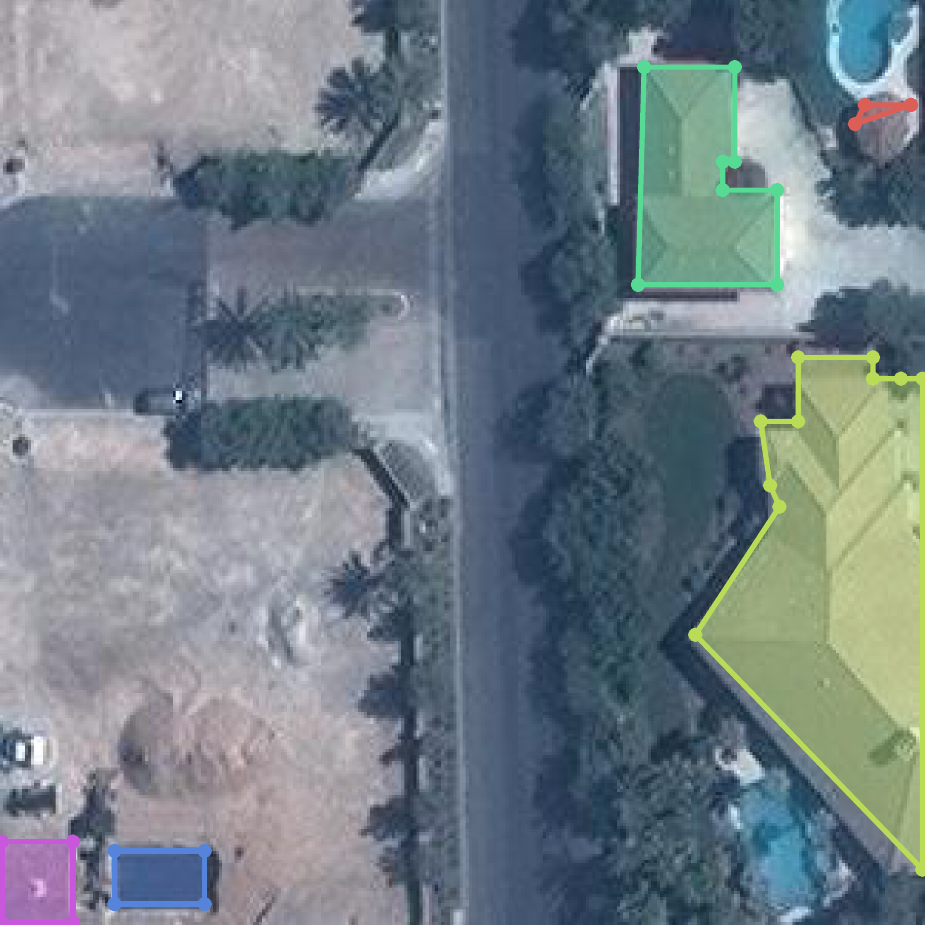}
\includegraphics[height=110pt,width=110pt]{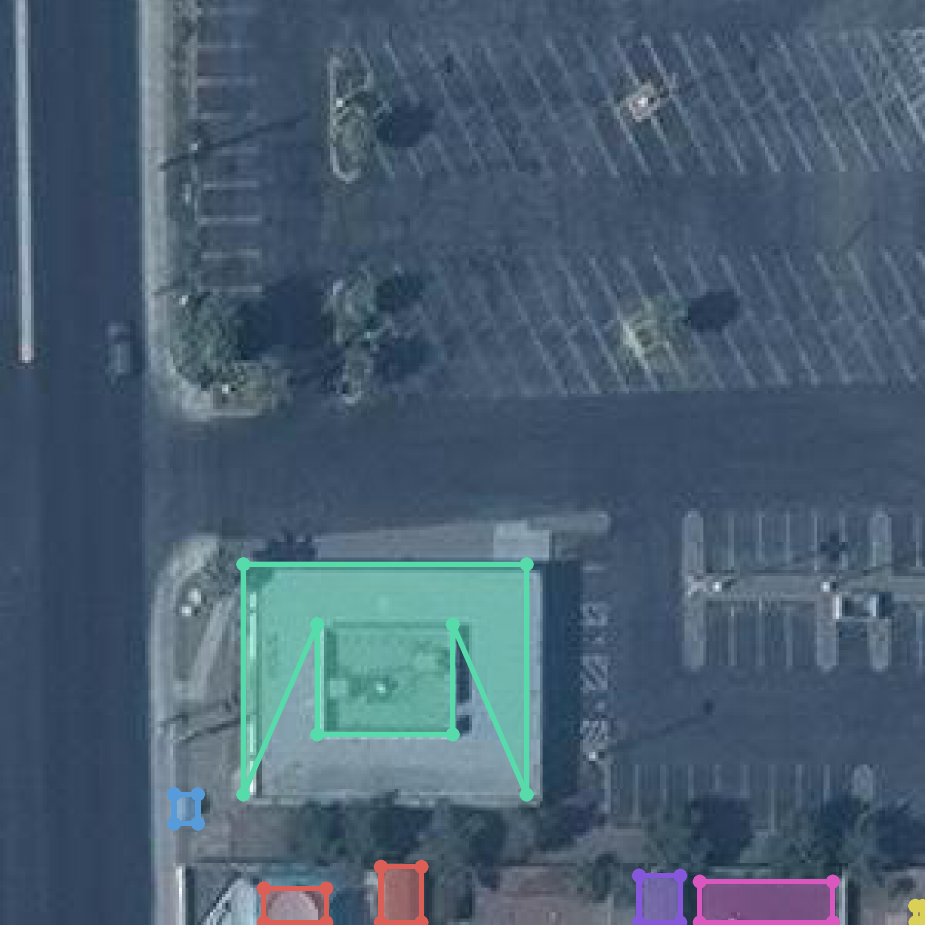}\\
\raisebox{52pt}{\rotatebox[origin=c]{90}{Frame-Field (ASM)}}
\includegraphics[height=110pt,width=110pt]{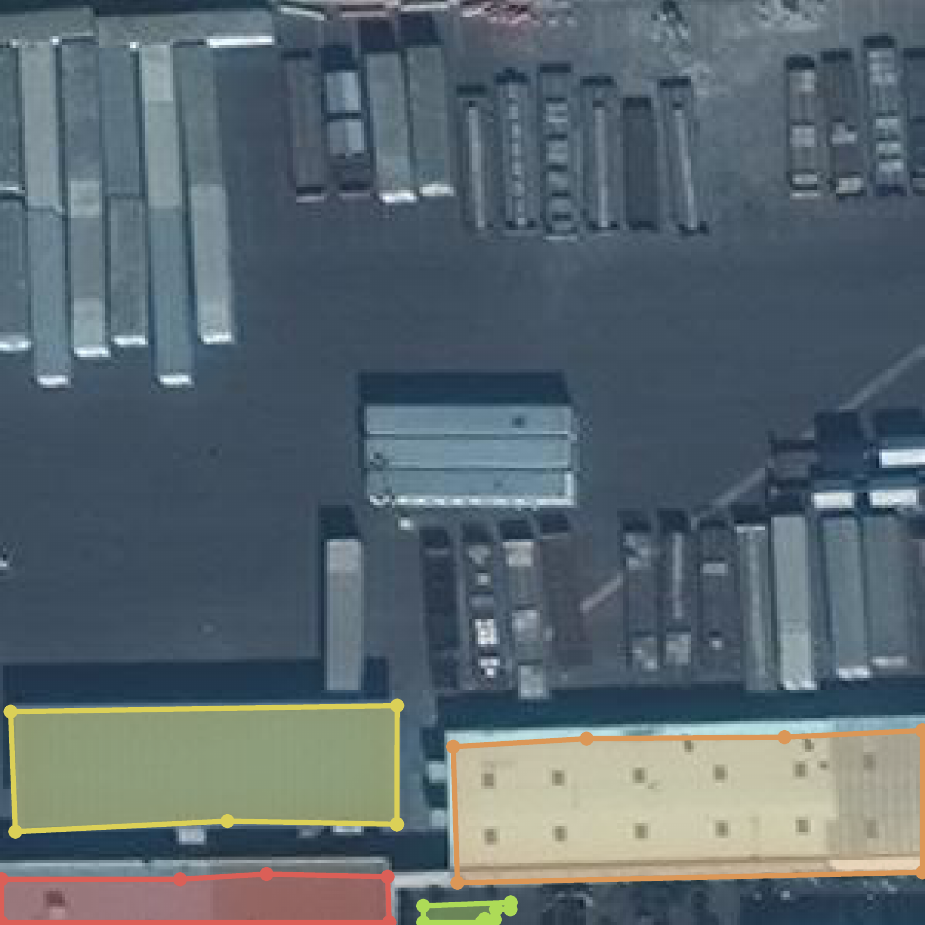}
\includegraphics[height=110pt,width=110pt]{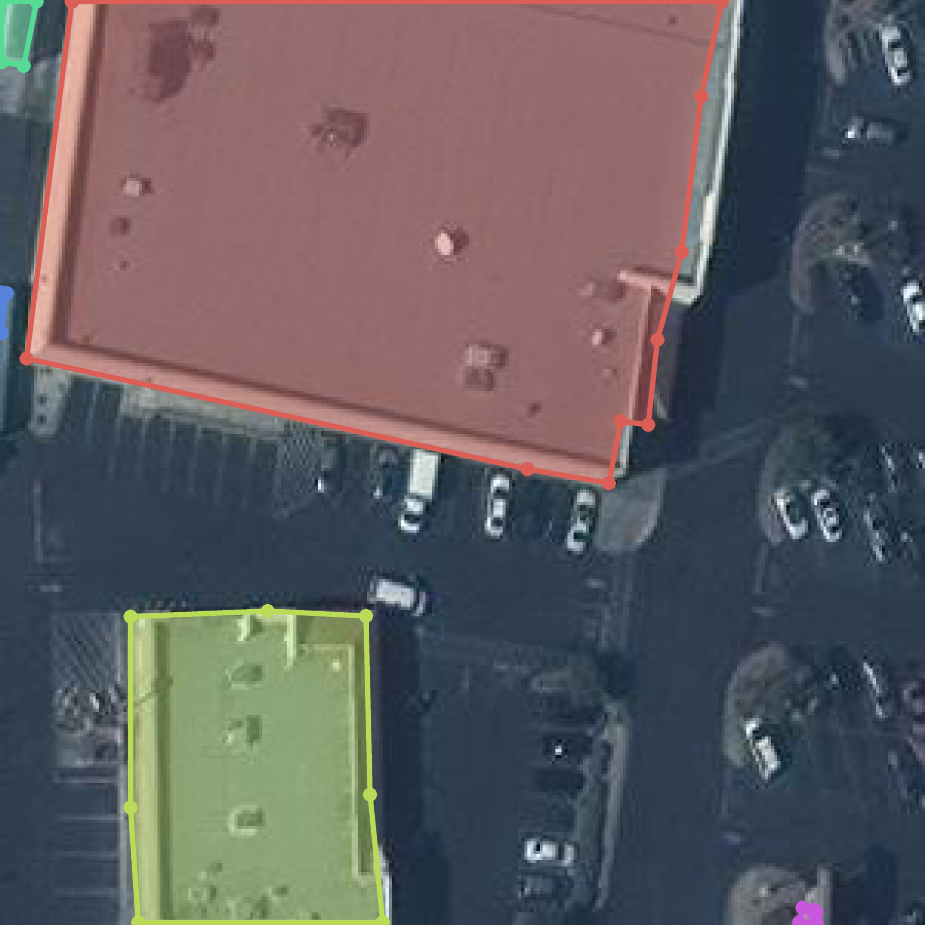}
\includegraphics[height=110pt,width=110pt]{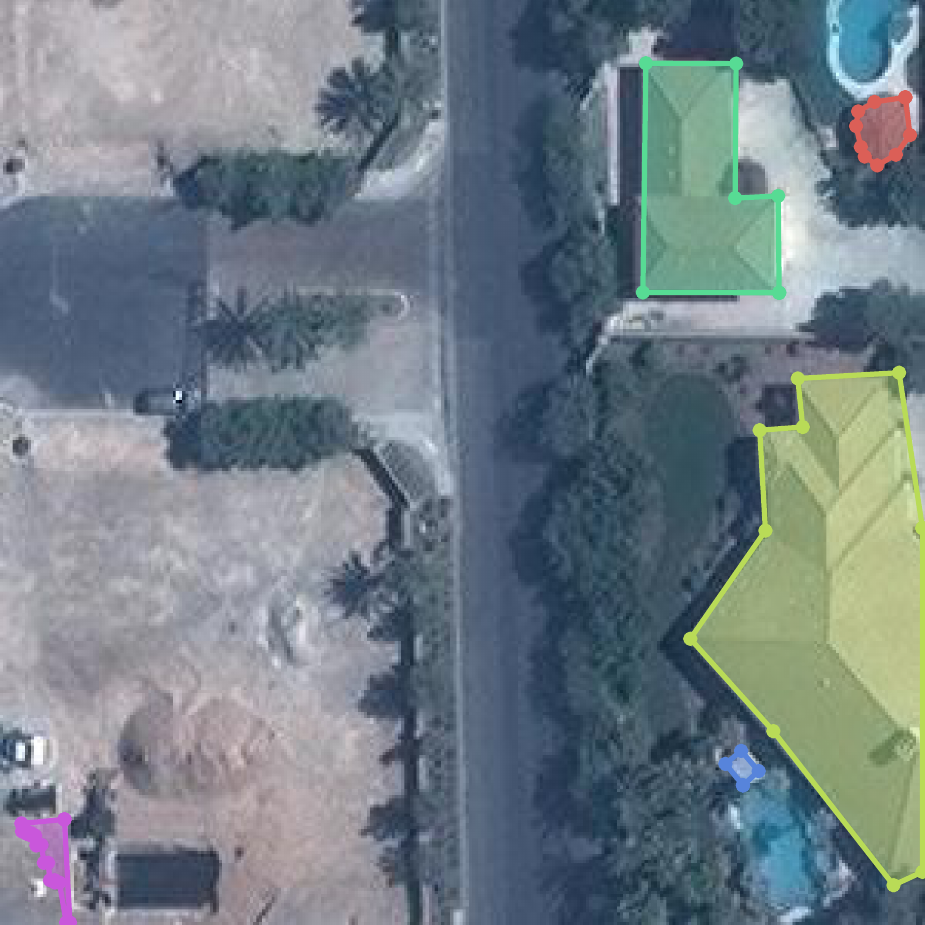}
\includegraphics[height=110pt,width=110pt]{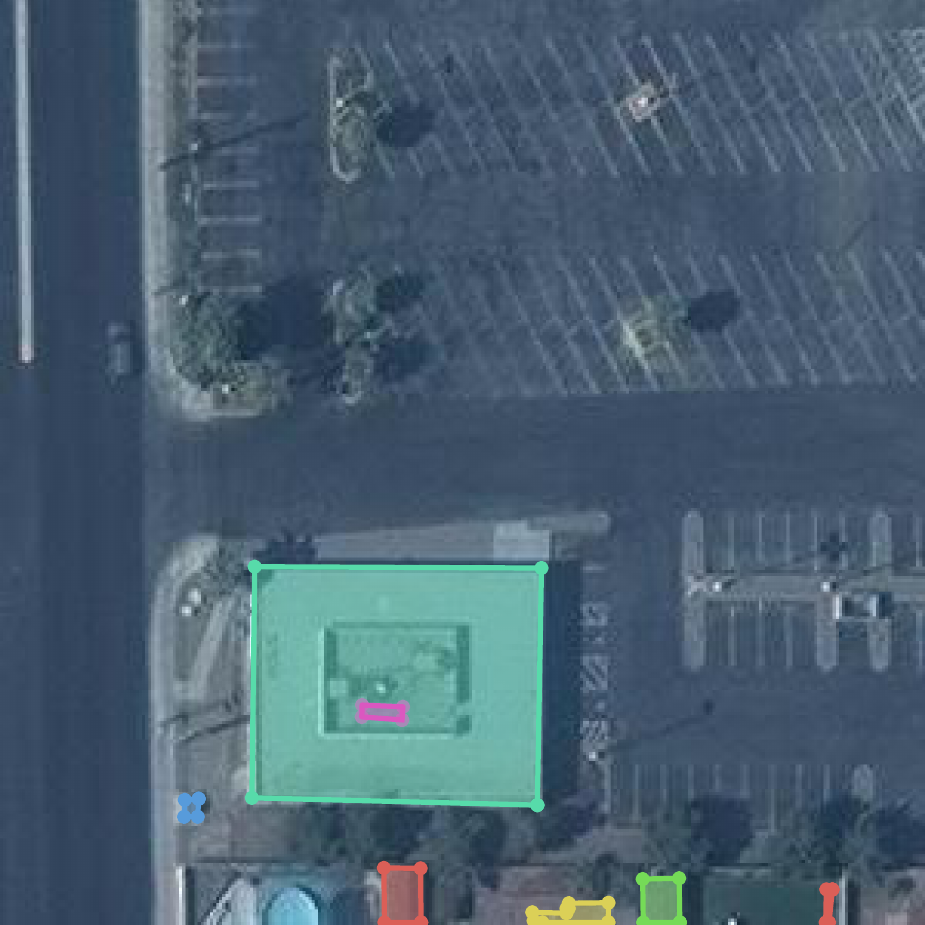}\\
\raisebox{50pt}{\rotatebox[origin=c]{90}{PolyWorld}}
\includegraphics[height=110pt,width=110pt]{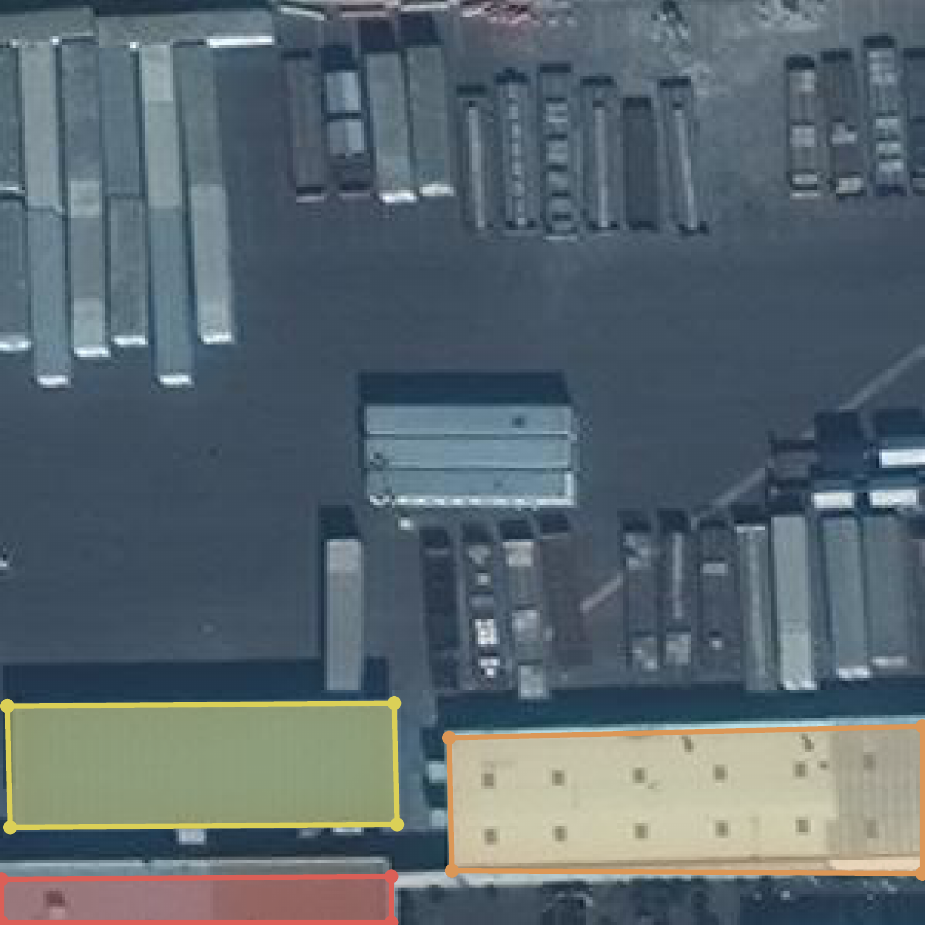}
\includegraphics[height=110pt,width=110pt]{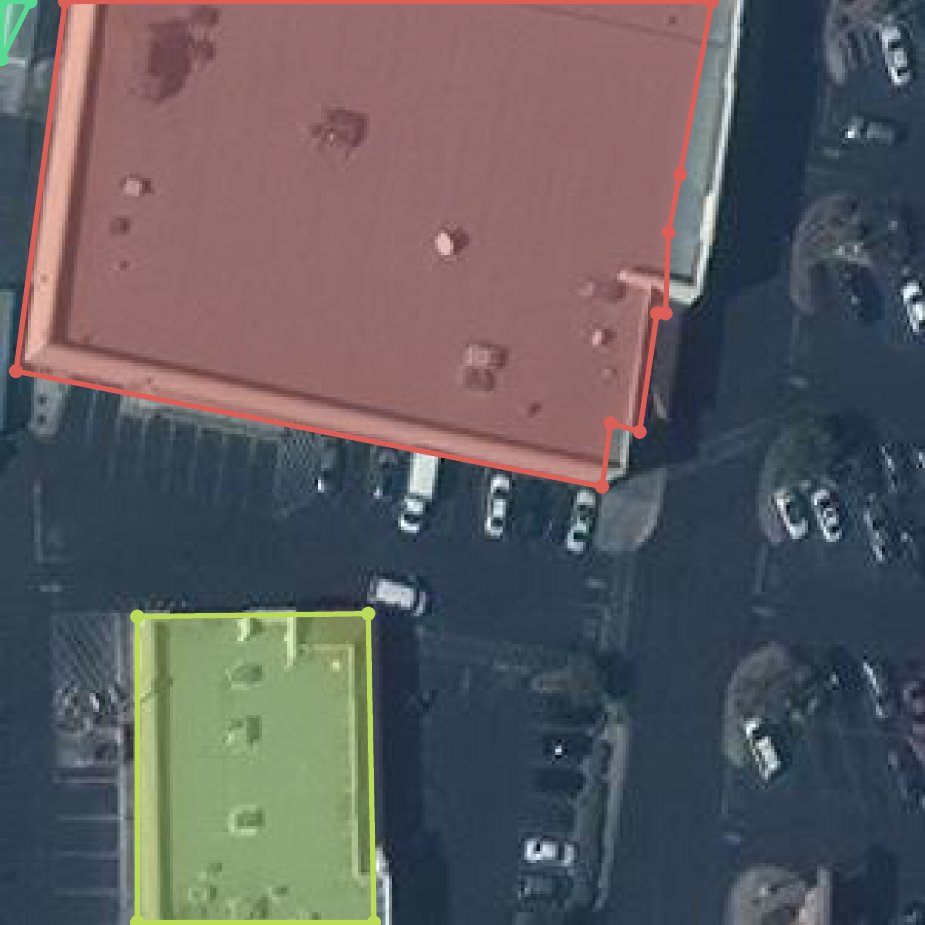}
\includegraphics[height=110pt,width=110pt]{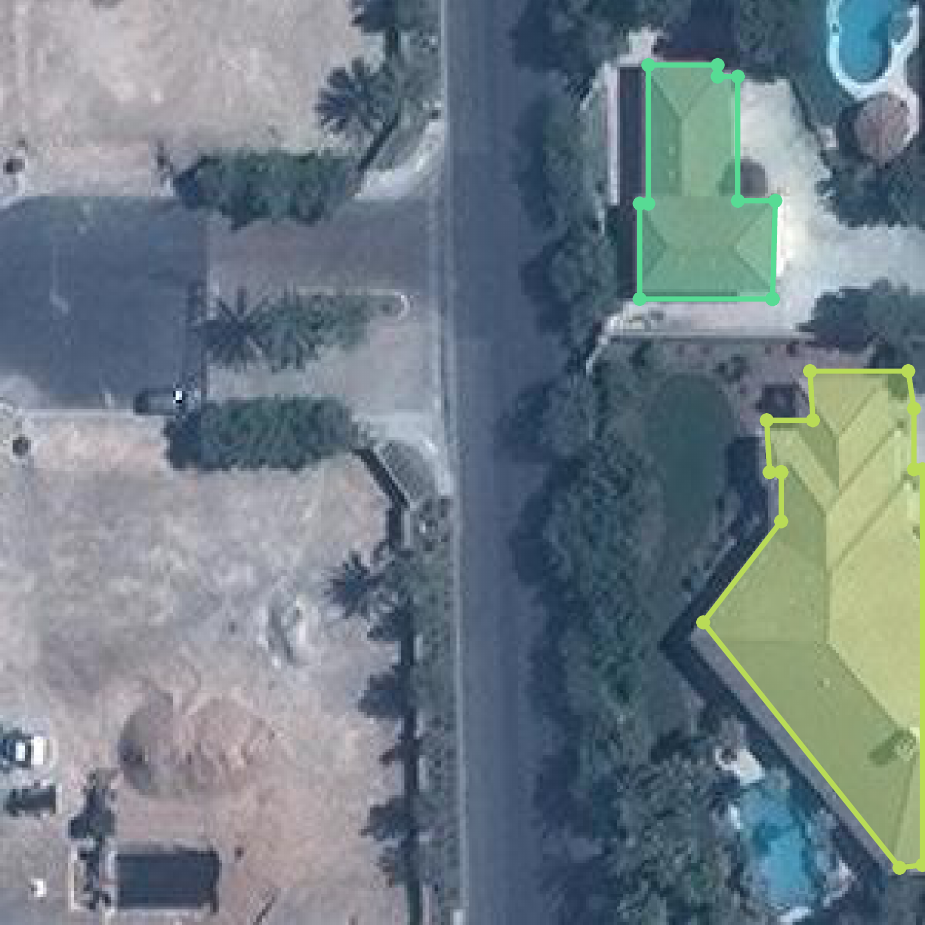}
\includegraphics[height=110pt,width=110pt]{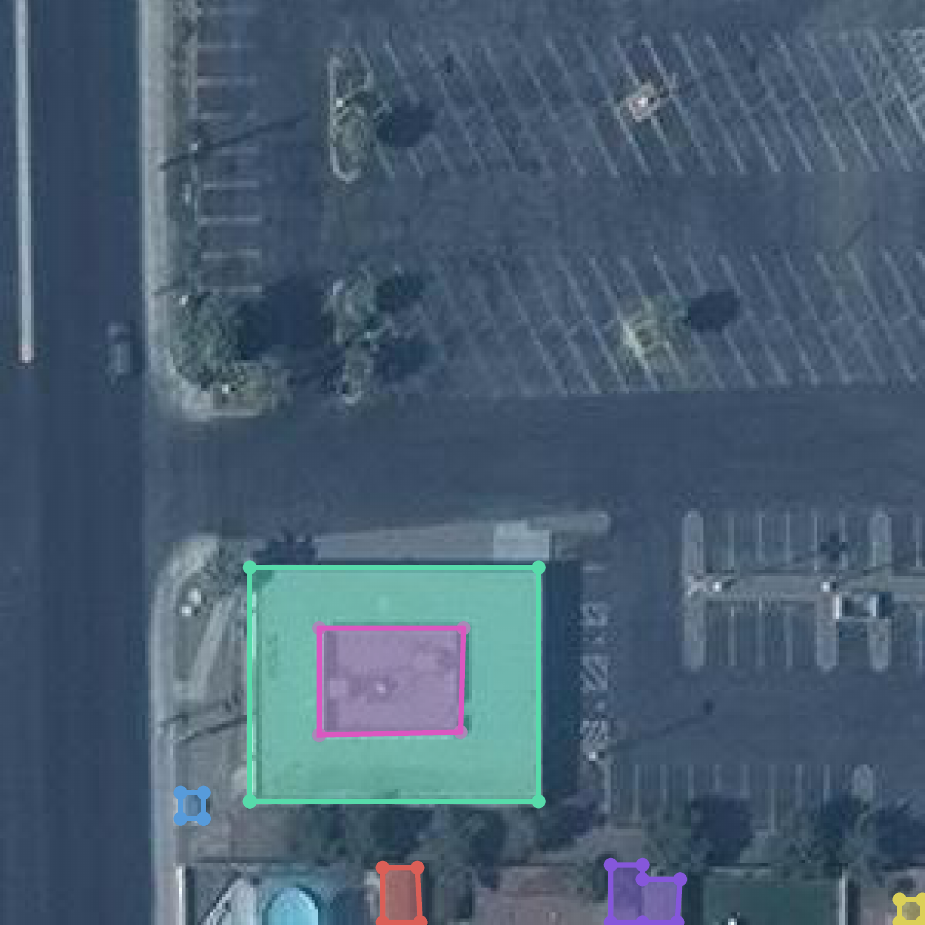}\\
\raisebox{50pt}{\rotatebox[origin=c]{90}{Ours}}
\includegraphics[height=110pt,width=110pt]{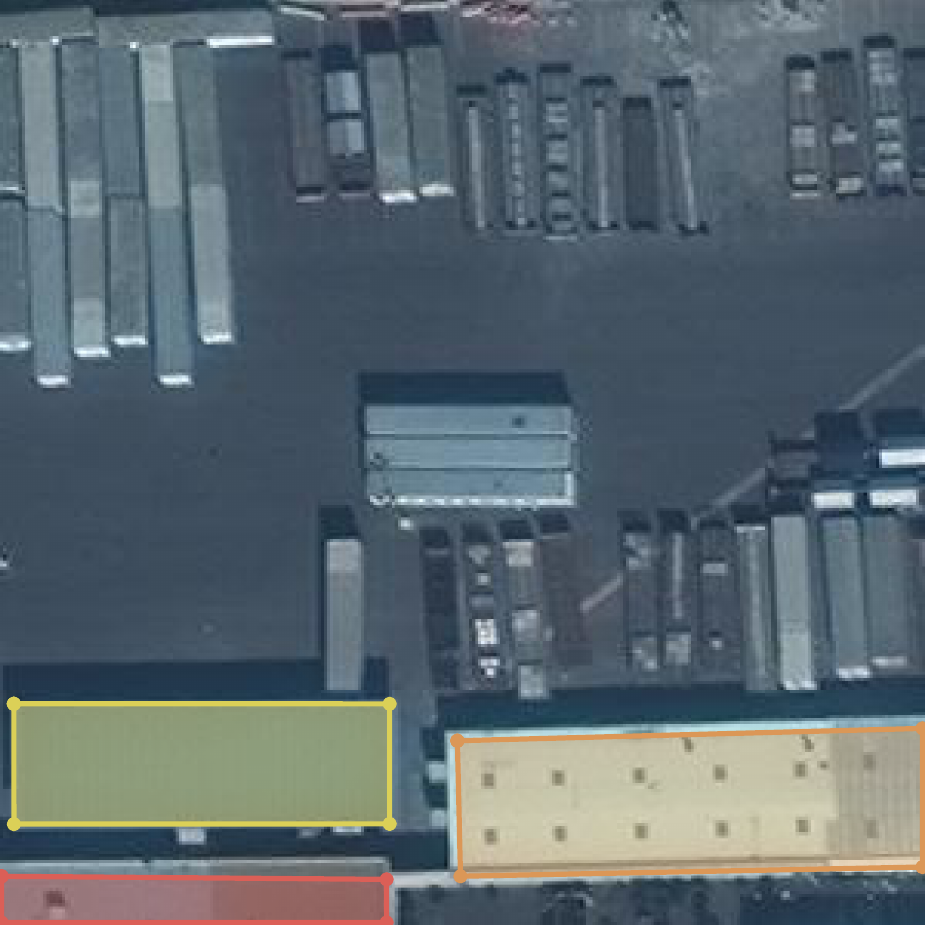}
\includegraphics[height=110pt,width=110pt]{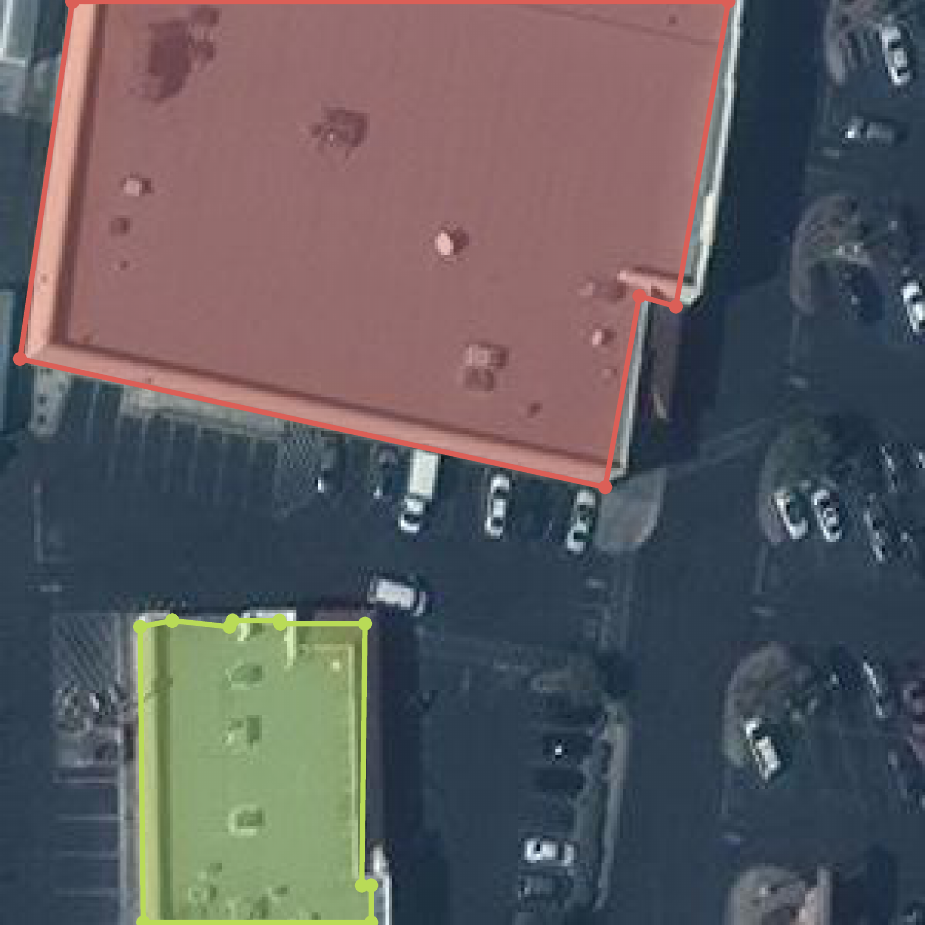}
\includegraphics[height=110pt,width=110pt]{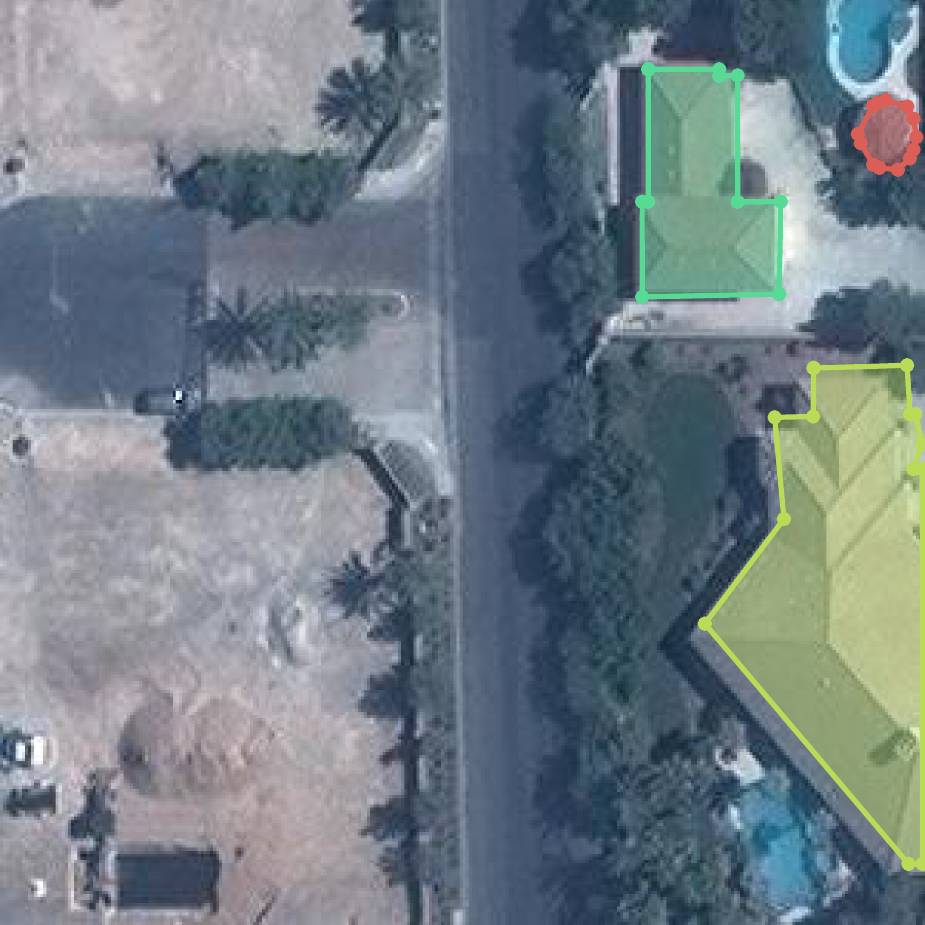}
\includegraphics[height=110pt,width=110pt]{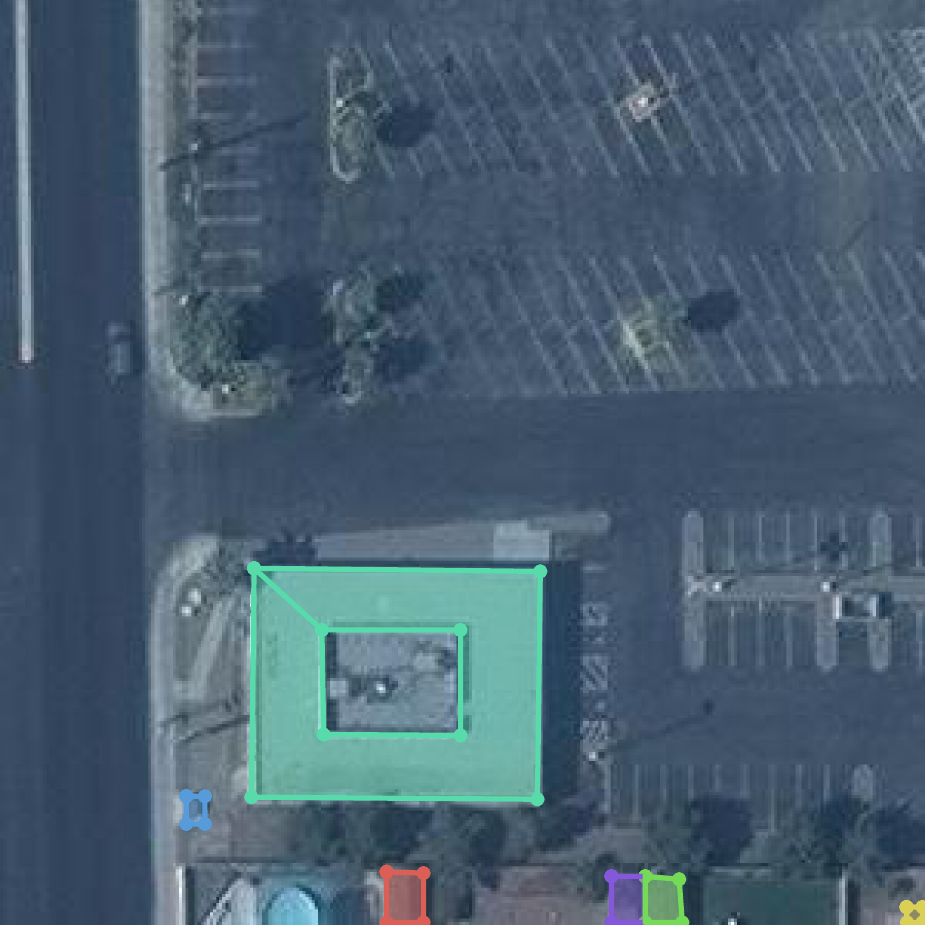}\\
\raisebox{50pt}{\rotatebox[origin=c]{90}{Ground truth}}
\includegraphics[height=110pt,width=110pt]{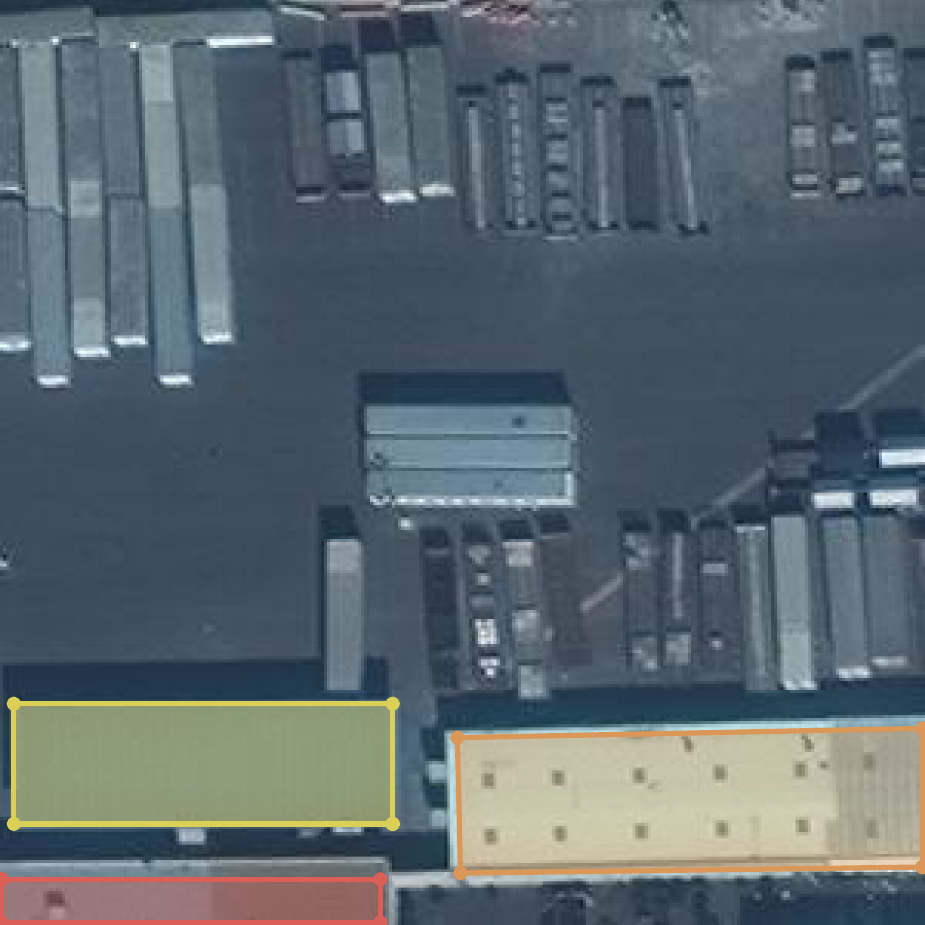}
\includegraphics[height=110pt,width=110pt]{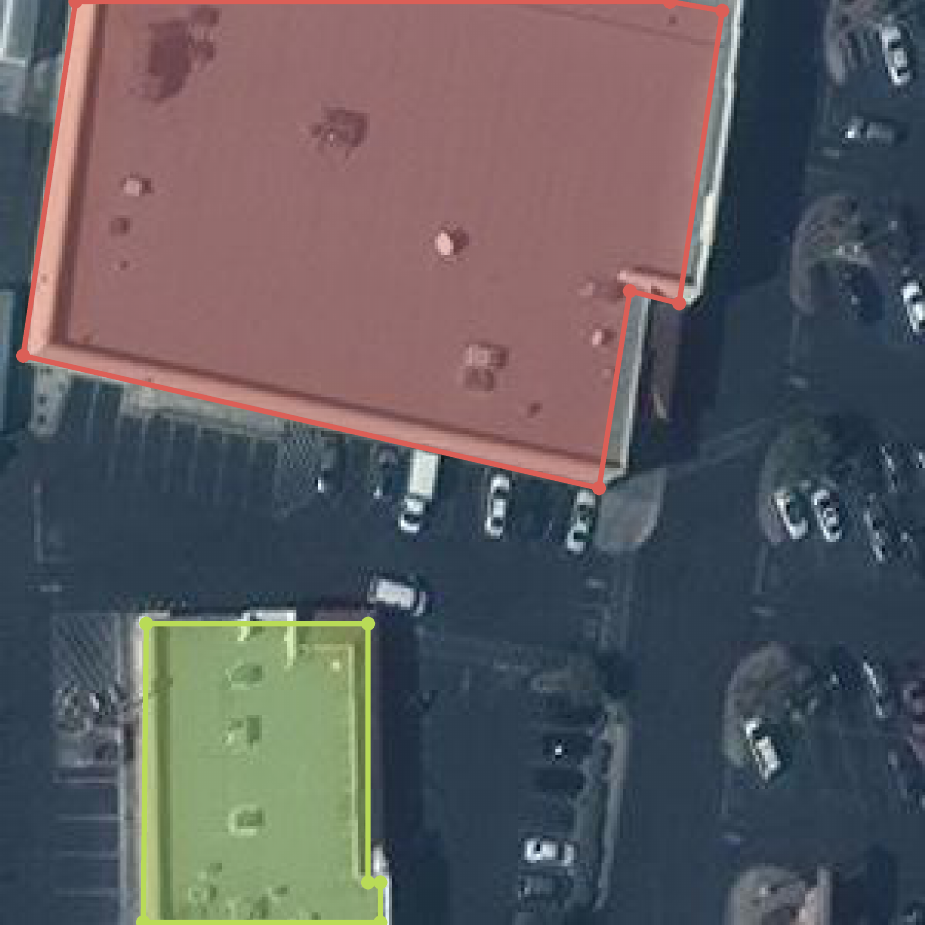}
\includegraphics[height=110pt,width=110pt]{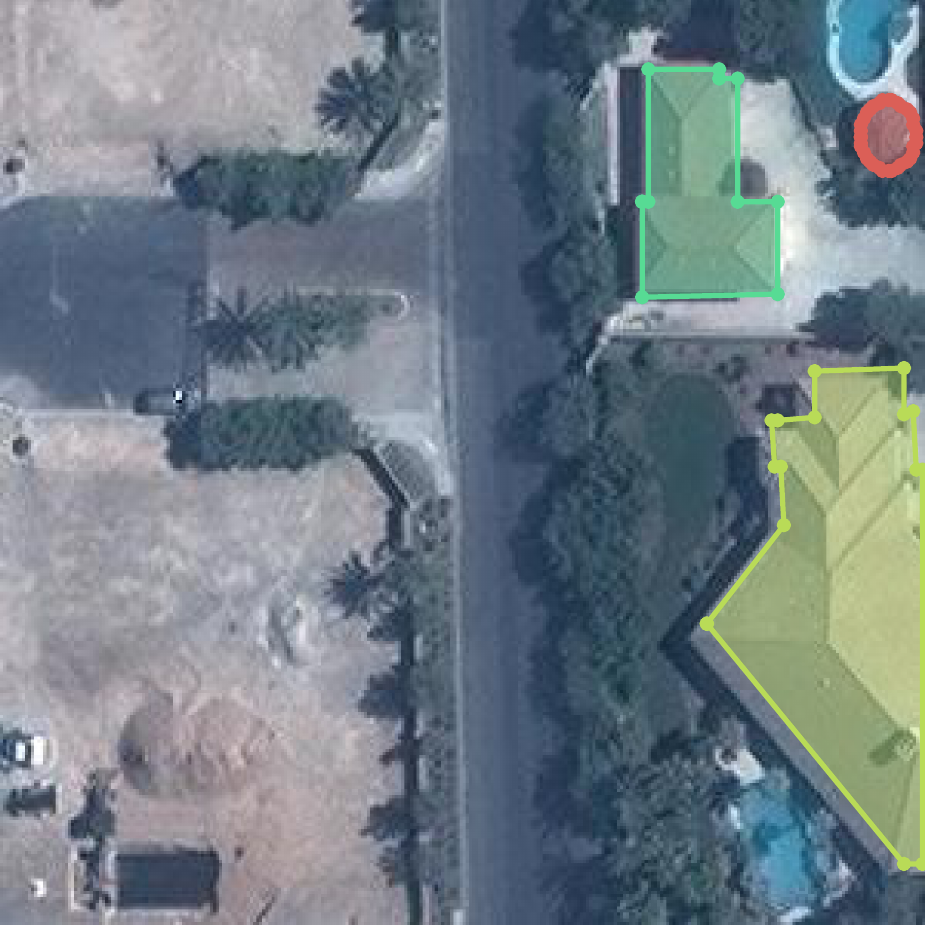}
\includegraphics[height=110pt,width=110pt]{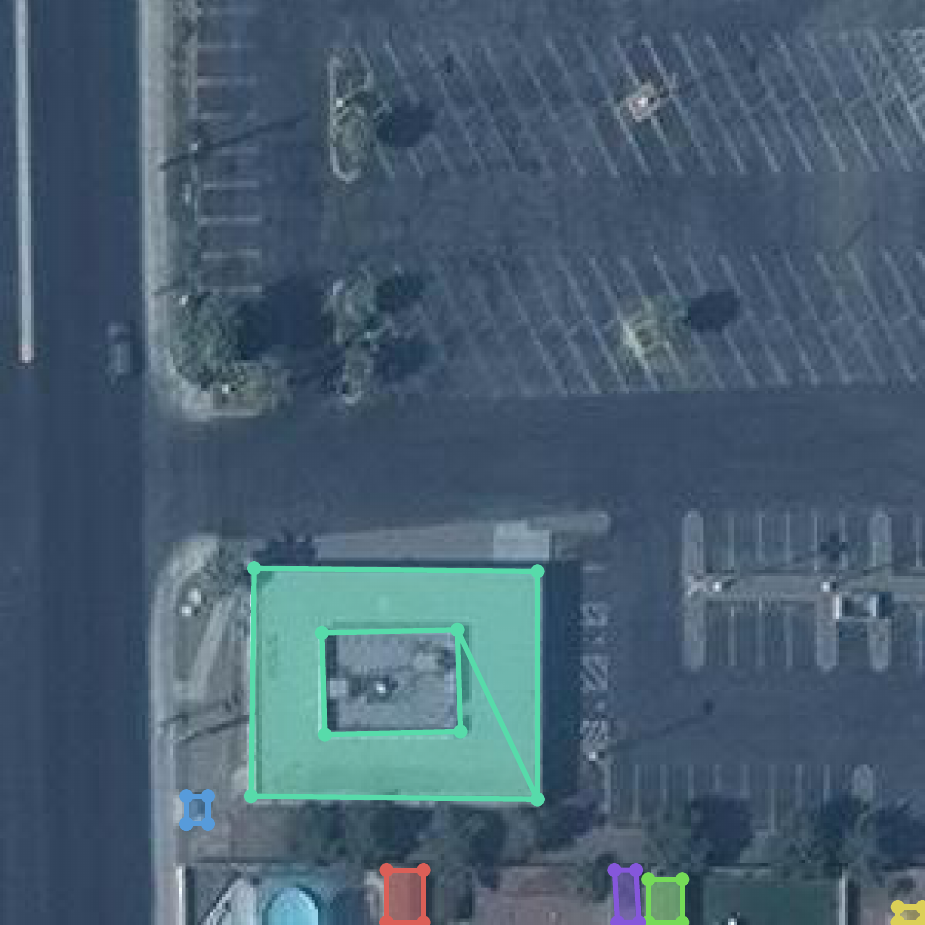}\\
\end{tabular}
\caption{Example of qualitative polygonal mapping of buildings results of AICrowd dataset. Building instances are marked with colors. From the top row to the bottom row: PolyMapper~\citep{Li2019topo}, Frame-Field~\citep{Girard2020polygonal} with ASM as polygonization and UNetResNet101 as backbone, PolyWorld~\citep{zorzi2021polyworld}, our results and ground truth.}
\label{fig:vis_crowdai}
\end{figure*}

Fig.~\ref{fig:vis_crowdai} shows the visual comparison of qualitative results. Each building instance is represented by a different colored polygon. Compared with the mask-related approach Frame-Field~\citep{Girard2020polygonal}, our method gets building polygons with fewer but more precise vertices. The PolyMapper~\citep{Li2019topo} and PolyWorld~\citep{zorzi2021polyworld} approaches are based on points inference. They both tend to predict relatively regular simple polygons. However, they are not good at dealing with buildings of complex structures. For instance, they failed in extracting the building marked in green hollow polygon in the image from the last column. While our method could both handle buildings with holes and get compact polygon representation as well. 

\begin{table*}
\centering
\caption{Comparison of additional results on the AICrowd dataset. The best scores are in bold.}\label{tbl_crowdai_poly}
\begin{tabular}{|l|l|cccc|}
\toprule
Method & Backbone & $\rm{AP}^{boundary}\uparrow$ & PoLiS\(\downarrow\) & C-IoU\(\uparrow\) & IoU\(\uparrow\)\\
\hline
Mask R-CNN~\citep{crowdAI2018baseline}                        & ResNet101  & 15.4 & 3.454 & 50.1 & 61.3 \\
PolyMapper\footnotemark[1]~\citep{Li2019topo}                 & VGG16      & 22.6 & 2.215 & 67.5 & 77.6 \\
Frame-Field (ASM)\footnotemark[3]~\citep{Girard2020polygonal} & UResNet101 & 34.4 & 1.945 & 73.8 & 84.3 \\
PolyWorld~\citep{zorzi2021polyworld}                          & R2U-Net    & 50.0 & 0.962 & 88.3 & 91.2\\
\hline
\textbf{HiSup (ours)} &HRNetV2-W48  & \textbf{66.5} & \textbf{0.726} & \textbf{89.6} & \textbf{94.3}\\
\bottomrule
\end{tabular}
\end{table*}
\footnotetext[3]{ASM refers to the Active Skeleton Model polygonization algorithm with the marching squares as initialization in Frame-Field~\citep{Girard2020polygonal}. Here the parameter of simplification tolerance equals 1. The whole pretrained UResNet101 model is downloaded from \url{ https://github.com/Lydorn/Polygonization-by-Frame-Field-Learning}.}

\subsubsection{Results on Inria dataset}
For the Inria dataset, our method is compared with the other five methods. The top two methods (``Advanced Institute'' and ``Eugene Khvedchenya'') in Table~\ref{tbl_inria} take the first and second place on the public leaderboard of Inria dataset website. The ICT-Net~\citep{chatterjee2019on} that combines improved UNet with Dense blocks~\citep{huang2017densely} and SE blocks~\citep{hu2020sqeeze} also got IoU score higher than 80. These three mentioned methods are trained directly on semantic segmentation annotations from the original Inria dataset. They all get segmentation results only and need extra post\hyp{}processing for vectorization. In addition, we list results of Frame\hyp{}Field~\citep{Girard2020polygonal} and~\citet{zorzi2020machine}, who get vectorized results as ours.  

Compared with AICrowd dataset, Inria dataset contains lots of buildings with more complex shapes and structures. And the patterns of building distribution in different cities vary greatly. Table~\ref{tbl_inria} shows the quantitative results of IoU and Acc on the testing images. Compared to the state-of-the-art polygonal mapping of buildings approaches~\citep{Girard2020polygonal,zorzi2020machine}, our proposed method achieves better performance on both the metrics of IoU and Accuracy. 

\begin{figure*}
\begin{center}
\includegraphics[width=\linewidth]{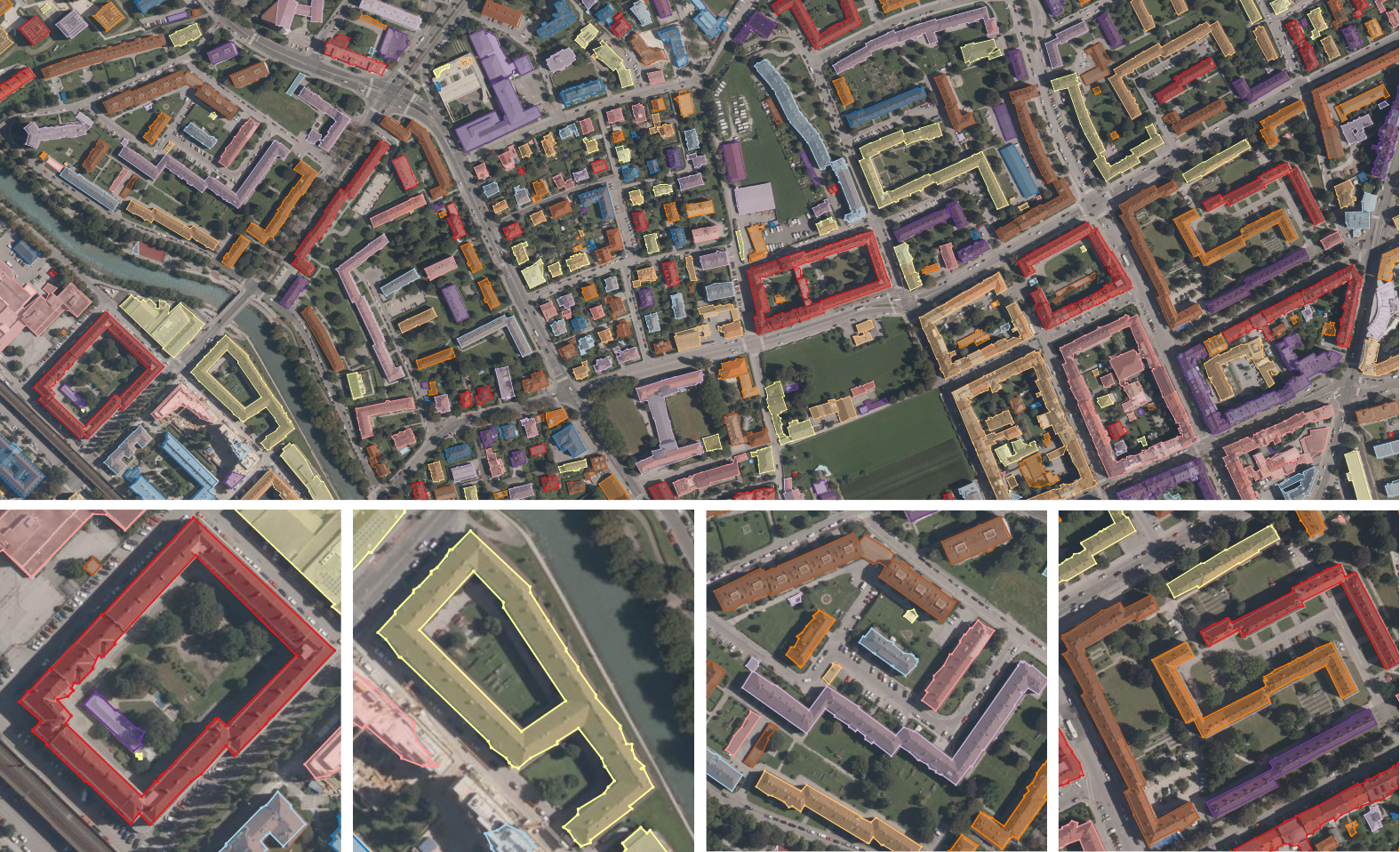}
\caption{Crop of polygonal mapping of buildings results by our method on a test image from the Inria dataset. Building instances are marked in different colored polygons. Four zoomed in views of polygonal mapping of buildings are provided as typical.}
\label{fig:inria-crop-result}
\end{center}
\end{figure*}

Fig.~\ref{fig:inria-crop-result} shows the qualitative results conducted on an image from the test set of Inria dataset. We produce vectorized polygons that fit the buildings with various shapes.

\begin{table}
\centering
\caption{Comparison of evaluation results on the Inria dataset.}\label{tbl_inria}
\begin{tabular}{|l|l|l|}
\toprule
Method &IoU\(\uparrow\) &Acc.\(\uparrow\)\\
\hline
Advanced Institute\footnotemark[4] &81.91 &97.41\\
Eugene Khvedchenya\footnotemark[4] &81.06 &97.25\\
ICT-Net~\citep{chatterjee2019on} &80.32 &97.14\\
\citet{zorzi2020machine} &74.40 &96.10\\
Frame-field~\citep{Girard2020polygonal} &74.80 &95.96\footnotemark[4] \\
\hline
\textbf{HiSup (ours)} & 75.53 & 96.27 \\
\bottomrule
\end{tabular}
\end{table}
\footnotetext[4]{Reported on \url{ https://project.inria.fr/aerialimagelabeling/leaderboard/}.}

\subsection{Discussion}
\label{discussion}
We perform additional studies to further validate the effect of details of our method. 

\subsubsection{Effectiveness of mask and vertex attraction}
On the basis of the same segmentation masks, we compare our polygon generation algorithm with other methods to validate its effectiveness. We generate the set of segmentation masks by applying our method to images of the AICrowd small validation dataset. The traditional DP simplification~\citep{douglas1973al} algorithm, ASIP~\citep{Li2020appro}, and Frame-Field~\citep{Girard2020polygonal} polygonization are chosen as comparison. The tolerance parameter of DP simplification~\citep{douglas1973al} is set to 1. For the ASIP~\citep{Li2020appro} polygonization, we kept the parameter settings the same as in their released code. For the Frame-Field~\citep{Girard2020polygonal} polygonization, we adopt the Active skeleton Model of tolerance 1 with UResNet101 as backbone and marching square as initialization.

As demonstrated in Table~\ref{tbl_poly_gen}, in terms of the segmentation evaluation metrics, the segmentation mask obtained by our method gets a fairly high score of 79.3\% for the AP. The DP simplification~\citep{douglas1973al} algorithm gets the AP of 69.5\% which is comparable to most existing methods. Based on the accurate segmentation masks, the ASIP~\citep{Li2020appro} and Frame-Field~\citep{Girard2020polygonal} polygonization methods get AP of 70.5\% and 73.7\% respectively. The scores are both higher than the originally reported results as shown in Table~\ref{tbl_crowdai}. Such differences validate our view that the reversibility of building masks is determinant for the final polygonization performance. 

Despite the comparable performance, the polygonization processes of DP and ASIP come with a great loss compared to the original masks. The segmentation AP of the generated polygons drops by almost nine points compared with the masks. The \(\rm{AP}^{boundary}\) also drops more than 10 points. The introducing of frame field vectors by Frame-Field learning method~\citep{Girard2020polygonal} narrows the gap between masks and polygons a little. In contrast, our method has the minimum loss when vectorizing the original masks. The gap of mask AP score is less than one point. By matching with vertices, the polygons obtained by our method have more accurate shapes, which leads to an even higher \(\rm{AP}^{boundary}\) score than masks. Besides, our generated polygons get the minimum PoLiS score. 

\begin{table}
\centering
\caption{Comparison of our polygon generation with other approaches on the AICrowd small dataset. ``Segmentation mask'' refers to the predicted mask results by our proposed network model. ``DP poly'' refers to the traditional Douglas\hyp{}Peucker simplification algorithm. ``ASIP poly'' refers to the ASIP polygonization method. ``Frame-field poly'' refers to the ASM polygonization method from Frame Field with UResNet101. ``Our poly'' refers to our polygon generation algorithm described.}\label{tbl_poly_gen}
\resizebox{\linewidth}{!}{
\begin{tabular}{|l|ccc|}
\toprule
Method &AP\(\uparrow\) &\(\rm{AP}^{boundary}\uparrow\) & PoLiS\(\downarrow\)\\
\hline
Segmentation mask                              & 79.3 & 65.8 & -      \\
DP poly~\citep{douglas1973al}                  & 69.5 & 51.3 & 1.178  \\
ASIP poly~\citep{Li2020appro}                  & 70.5 & 54.5 & 1.160  \\
Frame-field poly~\citep{Girard2020polygonal}   & 73.7 & 57.9 & 0.984  \\
Our poly                                       & 78.8 & 66.3 & 0.737  \\

\bottomrule
\end{tabular}
}
\end{table}

\subsubsection{Effectiveness of hierarchical supervision}
We validate the effect of hierarchical supervision by replacing the AFM with edge and frame-field vectors while the remained architectures of the network model stay the same. We also conduct experiments to evaluate the designed Cross-Level Interactions by just predicting AFM as a separate output to the backbone without the proposed interactions of AFM embedding. The baseline model is set as predicting only masks and junctions from backbone features. All mentioned models are trained and tested on the small version of AICrowd dataset.

The resulting masks and polygons are evaluated by the AP from MS-COCO metrics. The generated junctions are measured by F1-score with the controlled distance threshold set to 5 pixels.

As shown in Table~\ref{tbl_afm}, simply predicting AFM representation with a new head helps to improve all the mask, junction and polygon results compared to baseline. When introducing the proposed Cross-Level Interactions with AFM embedding, the performance has been significantly improved. The result also shows that the AFM supervision improves more on mask localization performance and junction detection performance compared to edge supervision and frame-field vectors supervision. 
\begin{table*}
\caption{Ablation study of the proposed hierarchical supervision on the AICrowd small dataset. The comparisons are made by replacing the supervision of AFM with edge and frame-field map respectively. The ``AFM (head only)'' stands for the simplified version of our model that omits the proposed cross-level interactions.}
\label{tbl_afm}
\centering
\begin{tabular}{|ccccc|ccc|}
\toprule
Baseline & Edge &Frame-field & AFM (head only) & AFM &Mask (AP\(\uparrow\)) &Polygon (AP\(\uparrow\)) & Junction (F\(\uparrow\))\\
\hline
\Checkmark & -         & -         & -         & -         & 57.1 & 55.9 & 59.3 \\
\Checkmark &\Checkmark & -         & -         & -         & 57.7 & 56.4 & 60.9 \\
\Checkmark & -         &\Checkmark & -         & -         & 57.7 & 56.4 & 61.6 \\
\Checkmark & -         & -         &\Checkmark & -         & 57.6 & 56.5 & 61.7 \\
\Checkmark & -         & -         & -         &\Checkmark & 58.1 & 56.7 & 62.6 \\
\bottomrule
\end{tabular}
\end{table*}

\begin{table*}
\caption{Evaluation of our method with different backbones on the AICrowd dataset.}\label{tbl_backbone}
\centering
\begin{tabular}{|l|ccccccc|}
\toprule
Backbone &AP\(\uparrow\) &\(AP_{50}\uparrow\) &AR\(\uparrow\) &\(AR_{50}\uparrow\) &\(\rm{AP}^{boundary}\uparrow\) & PoLiS\(\downarrow\) &C-IoU\(\uparrow\) \\
\hline
UResNet101  & 75.8 & 90.8 & 77.7 & 91.7 & 61.6 & 0.925 & 88.2\\
\hline
HRNetV2-W18 & 71.4 & 89.6 & 73.7 & 90.4 & 52.6 & 1.116 & 85.2\\
\hline
HRNetV2-W32 & 76.2 & 91.7 & 78.5 & 92.1 & 61.2 & 0.872 & 88.1 \\
\bottomrule
\end{tabular}
\end{table*}

\subsubsection{Different backbones}
We test our method with different backbone network architectures. The classical UResNet101 model is chosen as a comparison for utilized by most of the previous methods. For the HRNet~\citep{Wang2021hrnet} serials, we adopt the lightweight models of HRNetV2-W18 and the HRNetV2-W32. Table~\ref{tbl_backbone} shows the quantitative evaluation results of different backbone settings on the AICrowd dataset. It can be seen that our method with UResNet101 gets 75.8\% of the AP, which outperforms currently available methods. The larger model with the HRNetV2-W32 gets better performance than the HRNetV2-W18 and UResNet101.

\section{Conclusion}
\label{conclusion}
This paper studies the problem of polygonal mapping of buildings in a novel viewpoint of \emph{mask reversibility}. By taking the hierarchical-level supervision signals from the bottom-level vertices to the high-level regional masks, we present a novel method, HiSup, to use the mid-level attraction fields of line segments as the most important linkage. The key component of Cross-Level Feature Interaction in our proposed HiSup method learns a unified feature embedding in both aspects of high-level semantics and the shape correctness of buildings, thereafter closes the gap between the segmentation masks and polygons of buildings in satellite imagery. In the experiments, we show that the proposed HiSup outperforms existing polygonal mapping of buildings methods on several challenging metrics on two public benchmarks of AICrowd~\citep{Mohanty2020deep} and Inria~\citep{maggi2017can}. The systematic ablation studies further justified our design choices in our HiSup. 

\bibliographystyle{cas-model2-names}

\bibliography{cas-refs}







\end{document}